# Deep Learning Based Sentiment Analysis of COVID-19 Vaccination Responses from Twitter Data

**Kazi Nabiul Alam[1], Md Shakib Khan[1], Abdur Rab Dhruba[1], Mohammad Monirujjaman Khan[1,*], Jehad F. Al-Amri[2], Mehedi Masud[3] and Majdi Rawashdeh[4]**

[1]Department of Electrical and Computer Engineering, North South University, Bashundhara, Dhaka-1229, Bangladesh

[2]Department of Information Technology, College of Computers and Information Technology, Taif University, P. O. Box 11099, Taif 21944, Saudi Arabia

[3]Department of Computer Science, College of Computers and Information Technology, Taif University, Taif, 21944, Saudi Arabia

[4]Department of Business Information Technology, Princess Sumaya University for Technology, P.O.Box 1438, Amman 11941, Jordan

[*]Corresponding Author: Mohammad Monirujjaman Khan. Email: monirujjaman.khan@nortsouth.edu

**Abstract:** This COVID-19 pandemic is so dreadful that it leads to severe anxiety, phobias, and complicated feelings or emotions. Even after vaccination against Coronavirus has been initiated, people's feelings have become more diverse and complex, and our goal is to understand and unravel their sentiments in this research using some 'Deep Learning' techniques. Social media is currently the best way to express feelings and emotions, and with the help of it, specifically Twitter, one can have a better idea of what's trending and what's going on in people's minds. Our motivation for this research is to understand the sentiment of people regarding the vaccination process, and their diverse thoughts regarding this. In this research, the timeline of the collected tweets was from December'21 to July'21, and contained tweets about the most common vaccines available recently from all across the world. The sentiments of people regarding vaccines of all sorts were assessed by using a Natural Language Processing (NLP) tool named Valence Aware Dictionary for sEntiment Reasoner (VADER). By initializing the sentiment polarities into 3 groups (positive, negative and neutral), the overall scenario was visualized here and our findings came out as 33.96% positive, 17.55% negative and 48.49% neutral responses. Besides, the timeline analysis shown in this research, as sentiments fluctuated over time between the mentioned timeline above. Recurrent Neural Network (RNN) oriented architecture such as Long Short Term Memory (LSTM and Bi-LSTM) is used to assess the performance of the predictive models, with LSTM achieving an accuracy of 90.59% and Bi-LSTM achieving an accuracy of 90.83%. Other performance metrics such as Precision, Recall, F-1 score, and Confusion matrix were also shown to validate our models and findings more effectively. This study will help everyone understand public opinion on the COVID-19 vaccines and impact the aim of eradicating the Coronavirus from our beautiful world.

## 1 Introduction

COVID-19 outbreak brought significant attention to the healthcare sector in recent times and it changed the entire concept of safety in every aspect of our lives. Social distancing is an effective method to reduce spreading Coronavirus, besides wearing masks, washing hands repeatedly, staying concerned about intimacy, such safety measures are very important these days. But these can only reduce spreading

Coronavirus but can’t eradicate it completely. Here, Vaccination came into broad light, the only parameter that can fight most effectively against SARS-CoV-2 and probably kill it. Rigorous tests have been conducted with the first mRNAvaccines to be introduced; more than 40,000 people have participated in a Pfizer vaccine trial and 30,000 in the Moderna vaccine trial. The average efficacy rate of those who received vaccines was around 94 percent for the prevention of COVID-19 disease. There were no deaths from the trials, and in both trials, the participants received vaccines. Another viral vector vaccine, i.e., Johnson & Johnson, which proved to be able to fight against COVID-19 virus to stimulate the recipient's immune response, early findings show a rate of effective action of >85% without serious adverse effects [1]. Vaccination procedures are going in full swing all across the world, in Fig.1. There might be some reasons and conflicts among region to region in case of urgency and economic barriers, which will be explained later in our paper, but we tried to show the actual data of the numbers and figures without biasness, whether the people are vaccinated or not.

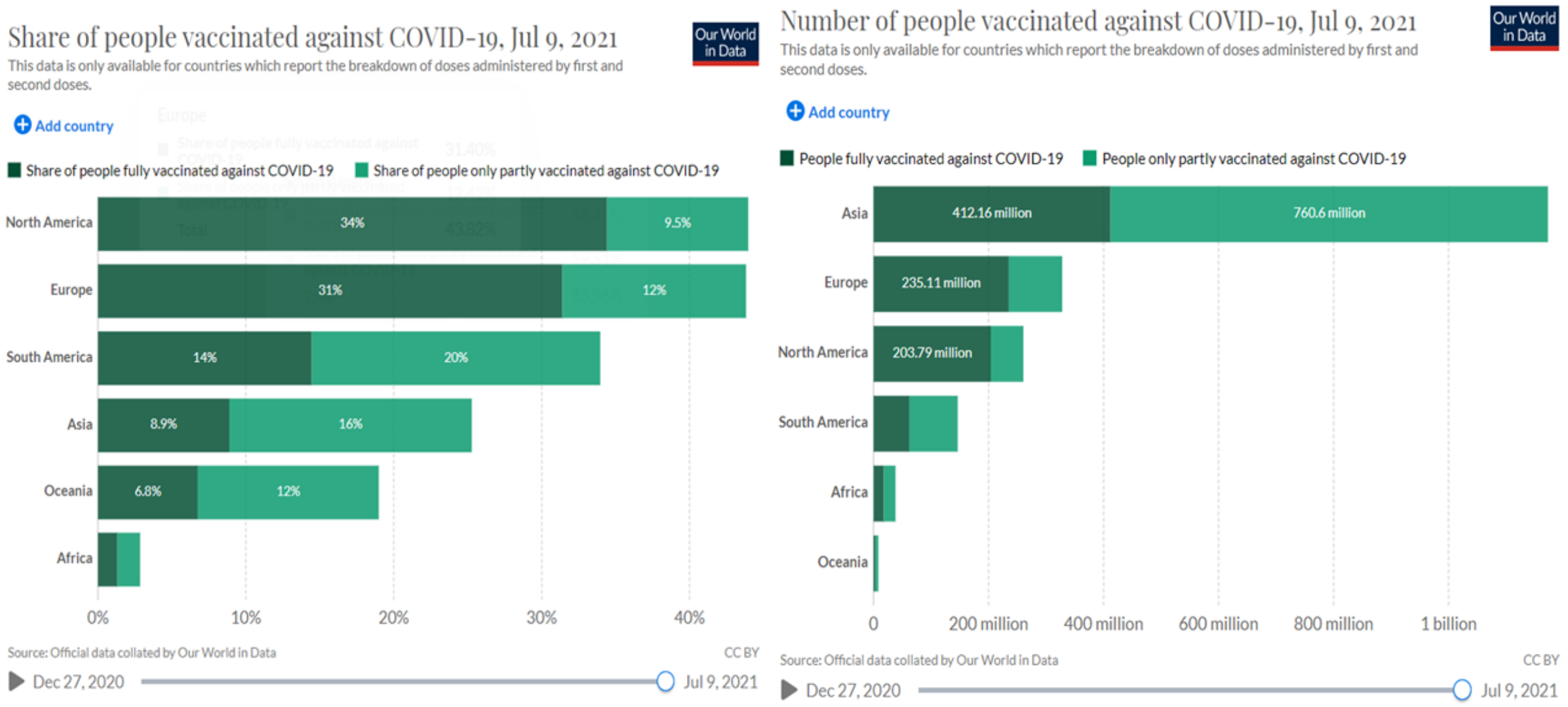

(a) (b)

**Figure 1**: Vaccination Progress in (a) Percentage and (b) Numbers around the World [2]

From this Figs. 1(a) and 1(b) above, it can be clearly observed that many of the populations in different continents are not vaccinated yet. Production of doses of these vaccines is a major concern, but the lack of willingness and interest is also an alarming factor, and it is a matter of great worry to the health scientists to find out the reasons. People have mixed feelings in this whole vaccination process from the very beginning, even though we heard such conflicts or questions from our family members. In different studies, researchers tried to understand the reason for such hesitancy. Some issues have recently been discussed in scientific articles, like the reasons people are thinking more than twice about getting vaccinated. Some reasons are: vaccines were invented so fast that there might not have been enough research on it, may cause cancer [3] or infertility, concerns about people getting 2nd dose or not, allergic reactions [4], blood clotting, legitimacy of the production industry, political belief, religious issues, social media and online trends [5], some conspiracy theories [6], [7], [8].

Several studies have been conducted to assess the feelings of people both before and after the advent of vaccinations. In [9], Snscrape was used in the period from 7 January, 2020 to 3 January 2021 to collect the historical tweets for vaccination against COVID-19. In all, 4,552,652 Twitter posts have been pulled. These tweets were produced by 1,566,590 users, with 1,012,419 numbers of hashtags and 2,258,307 terms of reference. They used VADER, a python lexicon and a rules-based sentiment analysis tool, which was developed to assess social media feelings on the basis of individual words and phrases, for assigning an amplitude of 'positive,' 'negative' or 'neutral' to each tweet. After extraction, they identified tweet vaccines and opinions and correlated their growth by period, geographic location, emerging topics, key phrases, postal engagement rate, and reports. The prevalence of positive and negative feelings was slightly different, with positive being the dominant direction and gaining large responses. Another research [10] found where, throughout the pandemic, tweets from the citizens of the United Kingdom and the United States were collected through Twitter 'Application Programming Interface (API)' and experiments conducted to answer three key vaccine questions: Positive, Negative and Neutral. Researchers performed relative sentiment analysis by VADER to get a dominant feeling of the citizens and introduced a modified approach, which can count the influence of the individual. In this way they were able to take the sentimental analysis a step further and explain some of the changes in the data. The three leading companies are identified: Pfizer, AstraZeneca and Johnson and Johnson, who are involved in research on vaccines and researchers [11] extract their Instagram posts from the start of vaccination and who receive data from users using their own hashtags. The company qualitative variations on manuscripts and visual characteristics, i.e. images categorization by transfer learning, are initially presented in this research. The 'Instaloader' was used to extract the images, and the image is classified with VGG-16, Inception V3 and ResNet50. Designing and conducting a controlled experiment confirms the accuracy ranking of the results algorithms and identifies the two best performing algorithms. Finally, the analysis of polarity of users' posts, using a Convolutional Neural Network (CNN), clearly shows a neutral to negative feeling with highly divisive user posts. This study [12] aims to perform a feeling analysis on the Twitter platform of both types of vaccines: Sinovac and Pfizer in Indonesia. Data was crawled and processed between October and November 2020 to understand the emotion. There were two types of datasets: Sinovac and Pfizer. In three classes, both data sets were marked manually: positive, negative, and neutral. After labeling and preparing data with 'Twitter Crawling', validating with 10-fold cross-validation, they performed Support Vector Machine (SVM), Naive Bayes and Random Forest to evaluate the performance and finally to get results with the proper labeling prediction. The authors of this study [13] collected information on the Philippines' feeling about efforts by the Philippine government using the Twitter Web. In order for the government to analyze its responses, NLP techniques were applied to understand the overall sentiment. The feelings were trained to categorize English and Philippine tweets as positive, negative, and neutral polemics via the data science tool 'RapidMiner' using the Naive Bayes model to classify them accurately. Another research [14] we found relevant, performed on Australian people sentiment collected from Twitter. This analysis aimed at extracting important issues and sentiments on Twitter Topics relates to COVID-19 vaccination by using machine-learning methods. They focus in particular on three factors: COVID-19 and its vaccination attitudes, the advocacy of COVID-19 infection control measures and COVID-19 control misconceptions and complaints. Between January and October of 2020, they collected 31,100 English tweets from twitter users in Australia containing COVID-19 related keywords. In particular, tweets were analyzed by illustrating high frequency textual data clouds and the interplay of word tokens. In order to identify the commonly mentioned subjects in a large tweet sample, they have created a Latent Dirichlet Allocation (LDA) model. Sentiment analysis was also performed to gain an idea of the overall feelings and emotions in Australia related to COVID-19. This research [15] proposed a machine learning framework based on bayesian optimization to detect COVID-19 and solve related issues on clinical perspective. An optimization approach [16] is developed that takes into account the individuals' isolation and social distance characteristics. The individual motivation grows by more than 85 percent with a rising proportion of home isolation, according to numerical data. A suggested game-theoretic incentive model is used to interpret the sustainability of the lockdown policy.

The above research indicates some highly satisfactory outcomes regarding COVID-19 vaccine reactions and their evaluation. Some of the research was related to sentiment related to pandemic tensions, and some were related to vaccination issues. But most of the research was found, prioritized region-wise or area wise, and mostly done with only Sentiment Analysis tools. Besides, many focused on particular countries and were specific to the vaccines of particular vaccine producing companies.

Our research analyzed the data of all the vaccines available, such as: Pfizer/BioNTech, Moderna, Oxford/AstraZeneca, Covaxin, Sputnik V, Sinopharm, and Sinovac, to understand the percentage of positive, negative and neutral sentiment percentage regarding vaccination against COVID-19. Our research is also focused on the timeline of such tweets to understand the sentiments, which is a novel contribution and very important finding because sentiment is very much related and changeable with the flow of time. Besides, our research showed with text inputs that our system can detect the sentiment of a sentence properly or not. The utmost objective  and contribution through this research is to give a clear idea about the emotions and thoughts of the general public regarding the vaccination process of COVID-19. This will help health researchers and policy makers take proper initiative to make these vaccines more unsuspicious and credulous and keep people safe and aware.

In the following sections, the technical components and their outcomes with analytical toolsare presented. Section 2 carries the methods and technical terminologies used to analyze the sentiments. In section 3, proper visualization tools are demonstrated, and the description of how these sentiments worked and what outcomes were found. Later the achievements of our research are discussed,and how it can be impactful and beneficial to mankind and how it can be further improved to bring betterment of the world. Our work is concluded in Section 4.

## 2 Methods and Materials

### *2.1 Full Outline of the Proposed System*

In this research, the dataset was collected from Kaggle [17] that contains different types of tweets related to the COVID-19 vaccine. After loading the dataset, checking the unique values, null values, and finishing preprocessing, data characters were detokenized to break the sentences into words and label them. Next, a sentiment column was added and calculated, which contains (Positive, Negative, and Neutral) using VADER. Then to evaluate the model, architectures of ‘Deep Learning’ called LSTM and Bidirectional LSTM was utilized to check the performance of the forecasting model. With the visualization tools, the outcomes were also visualized in graphs and text clouds. Sentiment classes, Timeline, Prevalent words and performance of the models are being clearly demonstrated and visualized. According to the proposed methodologies mentioned above, a full detailed outline of the system is demonstrated in Fig. 2.

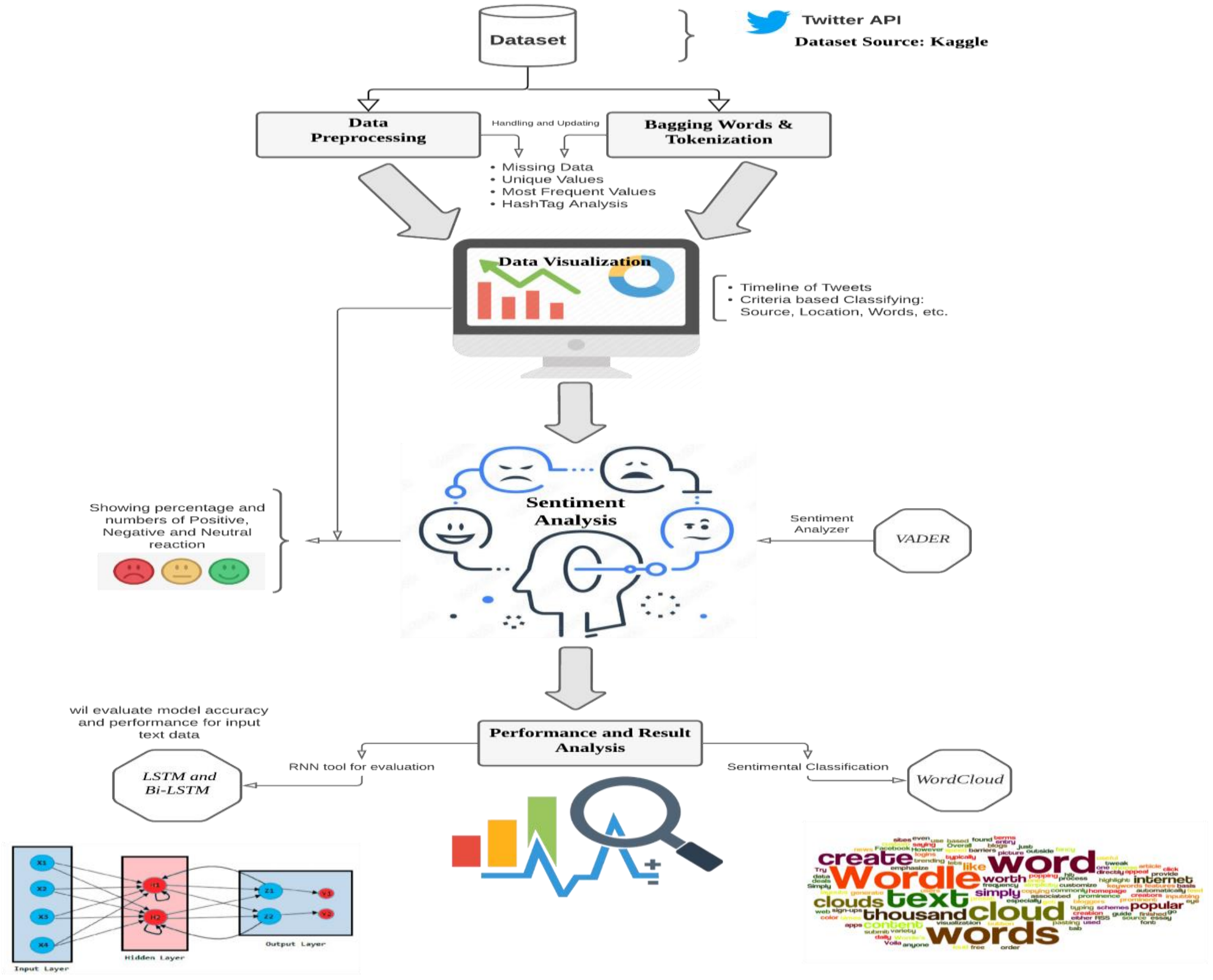

**Figure 2**: Outline of Sentiment Analysis Procedures

## *2.2 Data and Tools*

### *2.2.1 Dataset*

There are various datasets available for COVID-19 reactions after it broke out all over the globe, but there are limited numbers of dataset was found for 'Vaccination Reactions' related to COVID-19. From them, dataset named 'All COVID-19 Vaccines Tweets' from Kaggle [17] was chosen in our research wherethedata of almost all the renowned vaccines such as: Pfizer/BioNTech, Oxford/AstraZeneca, Moderna, Covaxin, Sputnik V, Sinopharm, Sinovac that are available and accepted to be implemented. Here, the dataset shape is (125906 by 16), including username, date, location, number of friends, retweets, hashtags, sources, etc. Our motivation behind choosing this dataset is that data contains all the vaccines available currently, and it helps us to generate precise and clear knowledge about vaccination reactions.

### *2.2.2 Implementation Tools*

Different kinds of Python packages and tools are used in this research. For analyzing and handling data, Google Colabotary is used to execute the codes, which have NVIDIA Tesla K80, high performing GPU, and built in environments. Natural Language Toolkit (NLTK) was used to handle and process the text data, and for visualization purposes Matplotlib, Seaborn, Wordcloud, such Python packages are used. Open source 'Machine Learning' platforms named Tensorflow and Keras were also implemented in running the deep neural models. Scikit-learn also played a vital role in evaluating the performance parameters.

## *2.3 Data Preprocessing, Handling and Tokenization*

The collected dataset was in the csv file format, which didnot have the processed data. Sothe data need to be processed first before applying any algorithm. Firstly, after dropping some unnecessary columns, all the URLs, emails from the tweets were removed. Then after taking out all the new line characters, alimenting all the double, single quotes, and all the punctuation signs were deleted. For this type of processing,all the tweets were tokenized before applying all the methods of removing those texts. After doing that, these were detokenized and converted as a NumPy array.

## *2.4 Sentiment Analyzer Tool (VADER)*

Valence Aware Dictionary for sentiment Reasoner (VADER), proposed by C.J. Hutto [18] in 2014, is a NLP-based sentiment analyzer, a pre-trained model that uses rules-based values tailored to the perceptions of social media expressions and which works well on texts from other fields. It analyzes a message text and appraises the intensity of this emotion like: positive, negative, and neutral. It is polarity and reflectance Intelligent Dictionary available in the NLTK package. In particular, it has impeccable performance in the area of social media text. Based on its comprehensive rules, VADER can perform a sentimental analysis of assorted lexical characteristics, shown in Fig. 3. VADER could therefore tackle the problems of indirect use of language groups and inherent action of sentiment.

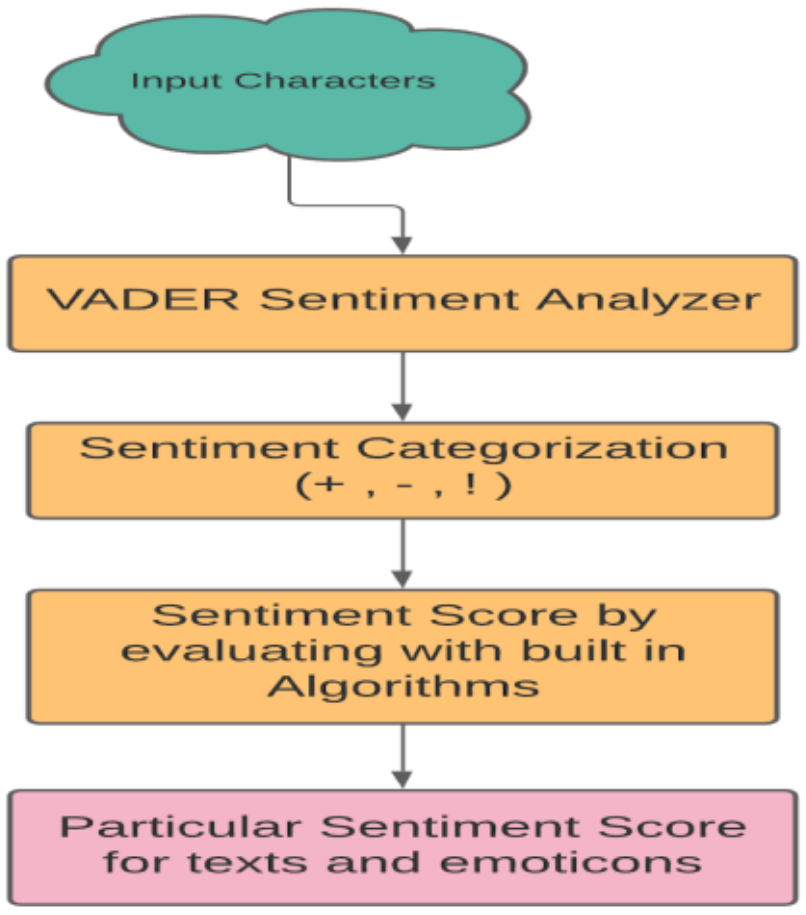

**Figure 3:** VADER diagram

As the valence values for each word in the lexicon, VADER provides a percentage for text ratios that crumble into a positive, negative or neutral category and sums up a probability value of 1. The compound score for sentiment analysis is the most frequently used measure where a float value in the interval [-1,+1] is a compound score, whose index is determined by adding in the lexicon the value values of each word, adapted according to rules and then standardized to its range. No training data is required. It can fully comprehend the vibe of a text which contains emotions, slang words, conjunctions, keywords, punctuation marks, etc., from different particular domains.

### *2.5 Data Visualization Tools*

Understanding patterns and correlations between numbers is at the heart of data visualization. Knowing structures, trends, and correlations in groupings of numbers is more important than understanding individual numbers. It may involve detection, measurement, and comparison from the user's perspective.It is strengthened by interactive techniques and information from many perspectives and multiple methodologies. In this work, various types of analysis were made to see how our data correlates with another visually. Different kinds of diagrams like Bar plots, Line graphs, WordCloudwere implemented to understand patterns between our datasets. To do visualization, we used lots of prebuilt libraries which are available in python. Tools like Matplotlib, Seaborn, and WordCloud help us to visualize what is hidden in the data.

### *2.6 Performance Evaluation Process*

A Recurrent Neural Network (RNN) based architecture called Long Short-Term Memory (LSTM) is being applied to evaluate models. Here, basic LSTM and Bidirectional LSTM (Bi-LSTM) are being implemented in our evaluation procedure. LSTM is a method of an adept RNN that addresses RNNs with additional cells, inputs, and outputs. Additional additives intuitively rectify the problems related to vanishing gradients, and gate activations are forgotten, enabling the gradients to pass through the network architecture without rapidly eroding. A Bidirectional LSTM or Bi-LSTM is a sequence processing model made up of two LSTMs. One takes the input forward and the other to the backward. Bi-LSTMs efficiently improve the volume of network information in order toimprove computational accuracy.

#### *2.6.1 LSTM*

An LSTM's control flow is similar toa recurrent neural network (RNN) shown in Fig. 4. As it moves forward, it processes data and passes it forward. The LSTM's cells have a variety of actions. The LSTM uses these processes to recall or forget information. The four gates that make up the LSTM architecture are sequential, input gate, forget gate, control gate, and output gate.

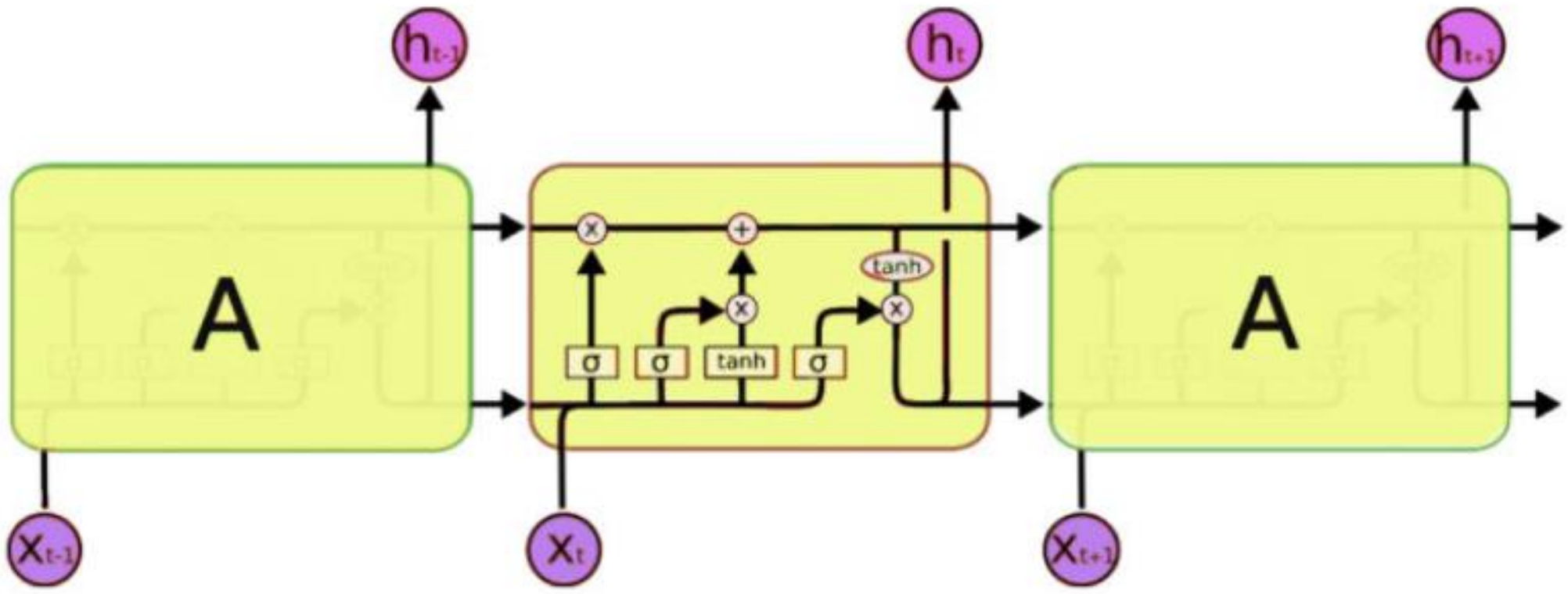

**Figure 4**: Schematic Diagram of LSTM [19]

A series of equations describes the gates of the LSTM [19]. Before attempting to describe the equation, it is necessary to first comprehend some of the variables used in these calculations. The sigmoid activation function is used, as well as the $W$ is weight matrices. The previous LSTM block's output is represented by $h_{t-1}$, and the preference for the corresponding gates is represented by$b_i$. Finally, $x_t$ is the existing timestamp's input. Here, the input gate $i_t$ described as in equation (1)

$$i_t = \sigma \, (W_i * \; [h_{t-1}, x_t] \; + b_i) \quad (1)$$

The data that can be given to the cell is chosen using this equation. The forget gate $f_t$ decides which data from the previous memory's input side should be ignored using the equation (2)

$$f_t = \sigma \, (W_i \; * \; [h_{t-1}, xt] \; + b_i) \quad (2)$$

Here in formula (3), tanh normalized the values into the range between -1 to 1, where $C$ is the candidate for cell state at the timestamp, controls the updating of the cell$(t)$.

$$\underline{C} = tanh \, (W_c \; * \; [h_{t-1}, \, x_t] \; + b_c)$$
$$C_t = f_t \; * \; C_{t-1} + i_t \; * \; \underline{C} \quad (3)$$

Output layer $(o_t)$ upgrades both the hidden layer $h_{t-1}$ as well as the output layer according to the formula (4)

$$o_t = \sigma \, (W_o \; * \; [h_{t-1}, x_t] \; + b_o)$$
$$h_t = o_t \; * \; tanh \, (C_t) \quad (4)$$

Here is our proposed LSTM configuration in Fig. 5.

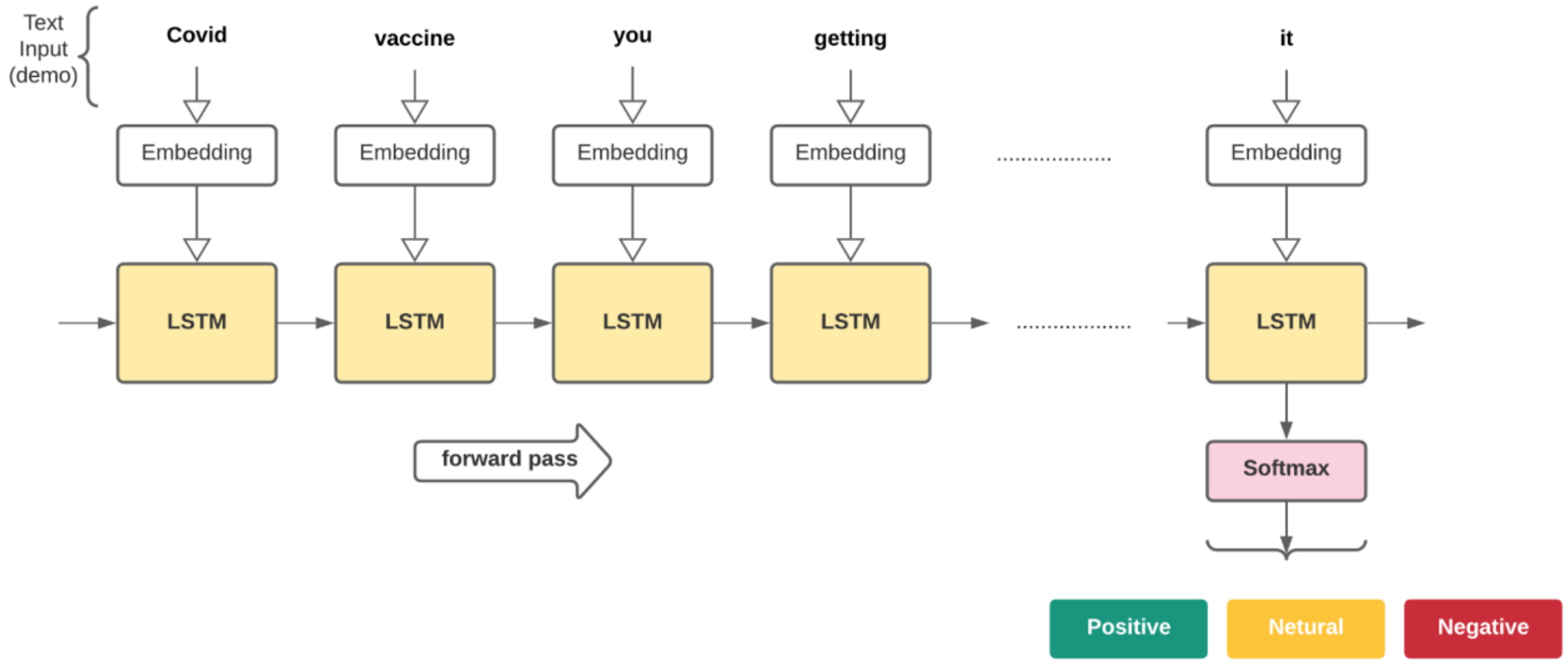

**Figure 5**: LSTM Architecture for Proposed Tasks

*2.6.2 Bi-LSTM*

Bi-LSTMs are inspired by bidirectional RNNs [20], that use two hidden layers to parse sequence input in both forward and backward paths, which is the motivation behind Bi-LSTMs. Bi-LSTMs combine two hidden layers to the same output layer.The configuration of an unfolded Bi-LSTM layer, which contains a forward and a backward LSTM layer, is described in this section and depicted in Fig. 6

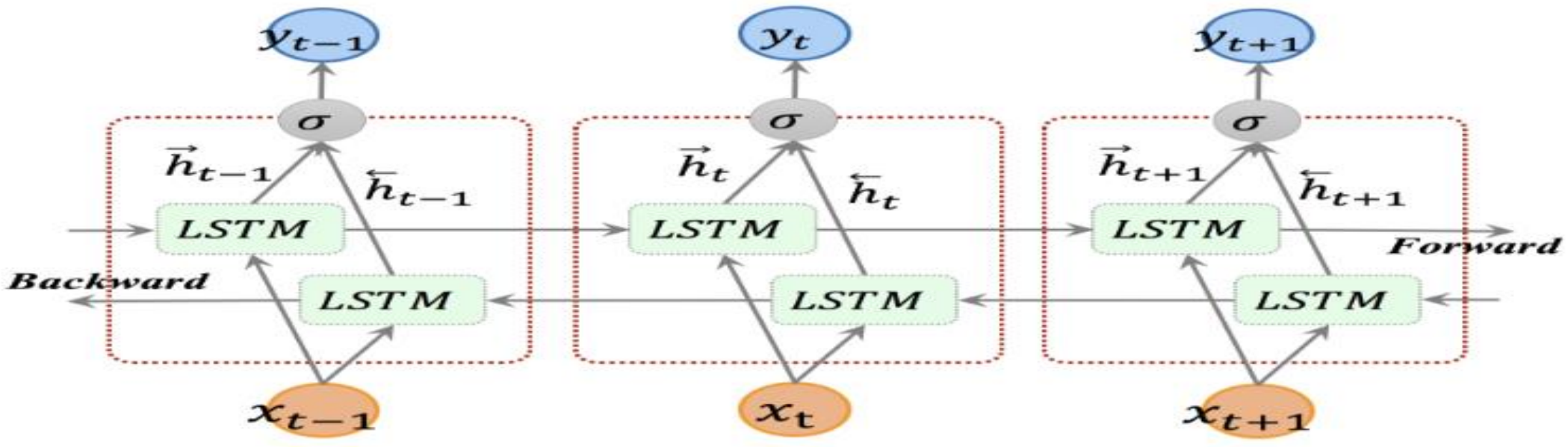

**Figure 6**: Bidirectional LSTM with Three Consecutive Steps [19]

The output sequence of forward layer $h^{\rightarrow}$is created repeatedly with positive sequence inputs from time $T{-}n$ to $T{-}1$, whereas the output sequence of backward layer $h^{\leftarrow}$ is measured by the reverse inputs from time $T{-}n$ to $T{-}1$. The basic LSTM equations (1) - (4) are used to calculate both the forward and backward layer outputs. The Bi-LSTM layer produces an output in vector form $Y_T$, in which each element is calculated by using the equation (5) below:

$$y_t = \sigma\ (\ h^{\rightarrow}\ , h^{\leftarrow}\ ) \qquad (5)$$

The two output sequences are combined using the $\sigma$ function. It could be a concatenating, summing, averaging, or multiplying function. A Bi-LSTM layer's final output can be represented as a vector in the same way that an LSTM layer's final output can, which is$[Y_T = Y_{T-n,......,}Y_{T-1}]$, in the last element, $Y_{T-1}$, is anticipated in the following iteration.

Proposed Bi-LSTM architecture is designed below in Fig. 7.

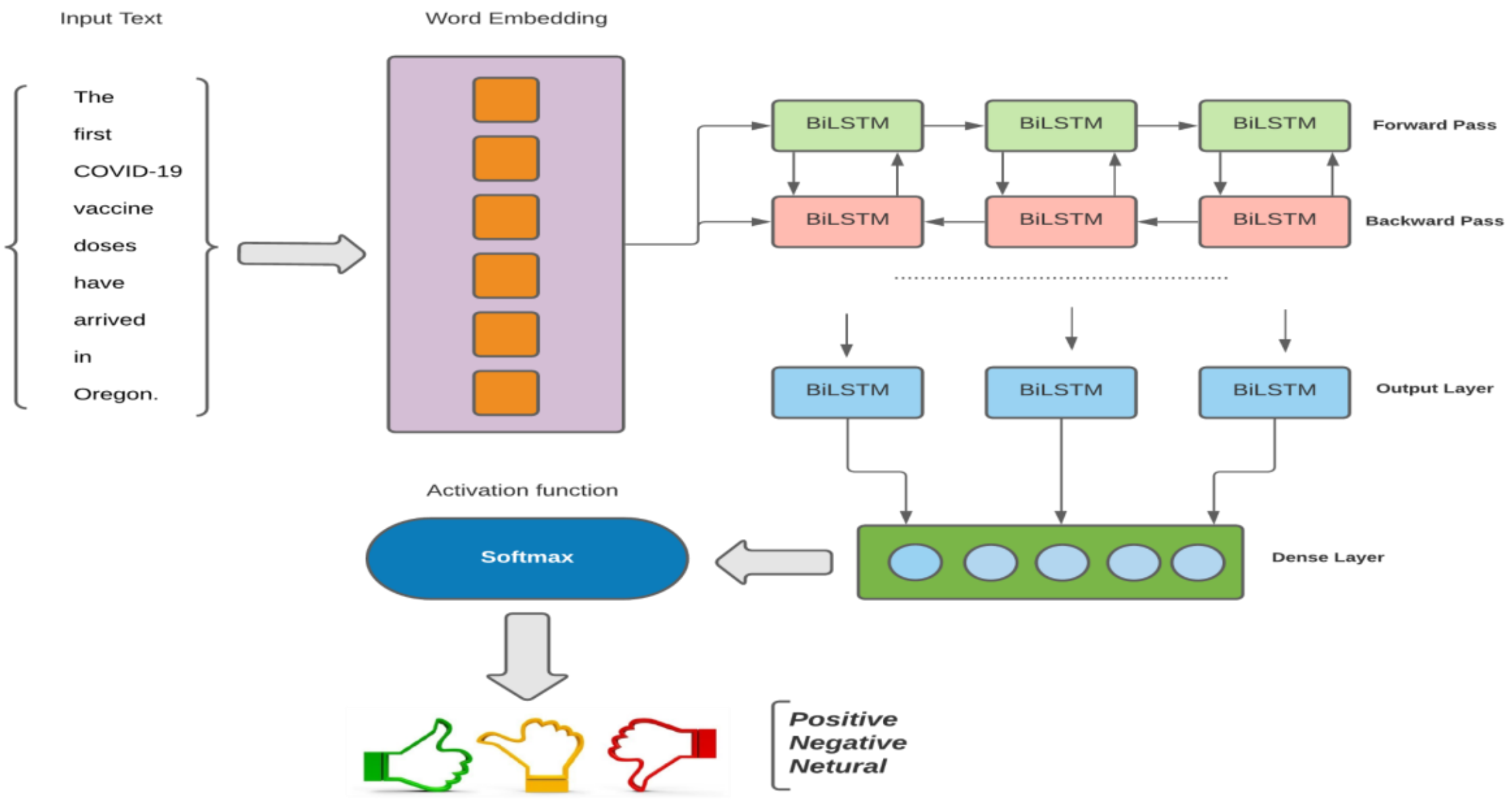

**Figure 7**: Bi-LSTM Architecture for Proposed Tasks

In following section, all the outcomes visualized and discussed with their analytical results.

## 3 Results and Analysis

Several outcomes were found and tried to dissect in our research. A brief discussion of such findings is presented in the following sections.

### *3.1 Overview of Outcomes*

Vaccines available today didn't come with ease. Researchers and Scientists tried hard to find out its killer antibody from the very earliest time of detecting Novel Coronavirus. Pfizer/BioNTech, Oxford/AstraZeneca, Moderna, Covaxin, Sputnik V, Sinopharm, and Sinovac are the most common and acceptable vaccines nowadays. And after its production and so many trials on different occasions, there was a timeframe to become available for the mass. So here timeline is important because sentiment and thought changes with time, especially in this pandemic situation. Another key thing to keep in mind is how efficiently our work can analyze the fact that people are still or ever interested in being vaccinated, how their psychology works, how they interpret their feelings, and its defense.

People's reactions vary from country to country found in this research. The reason behind is the stability of the government, faith in vaccine producing companies, nationalism, and other factors. Hashtag analysis is another important feature of Sentiment Analysis that we did in our experiment. The most important part was to classify the tweets in the form of ratio analysis to show the numbers and percentage of the reaction: Positive, negative, and neutral, as shown in our experiment. Then we validate our model using special sort of RNNs' such: LSTM and Bi-LSTM. Both architectures are pretty accurate, invalidating and give better accuracy in our model. We analyzed the relative metrics and accuracy-loss and other equivalent indicators of performance evaluation methods and showed them graphically. Using the LSTM and Bi-LSTM architecture, we showed how well our model could predict particular characters' inputs (tweets). Wordcloud is a frequently used visualization tool in Sentiment Analysis tasks that we used to show words are categorized as 3 different types of sentiments to understand the psychological analysis from these tweets.

### *3.2 Tweets according to User Locations and Sources*

Tweets are being collected from different sources from different parts of the globe; we analyzed them with facts and figures below. In Fig. 8, the sources of such vaccine reactions are showed asTwitter was the most used platform for delivering these thoughts in the COVID-19 situation. And as the data contains tweets from all around the world, we tried to show in Fig. 9 some major regions where tweets came from.

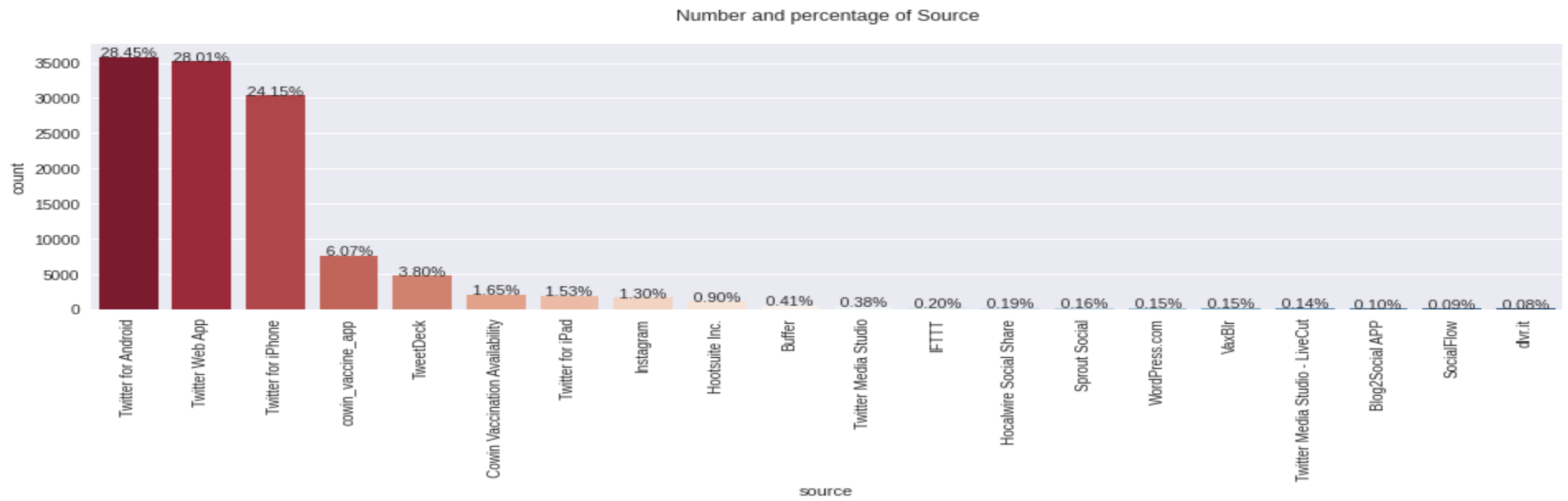

**Figure 8**: Sources of Vaccine Reactions

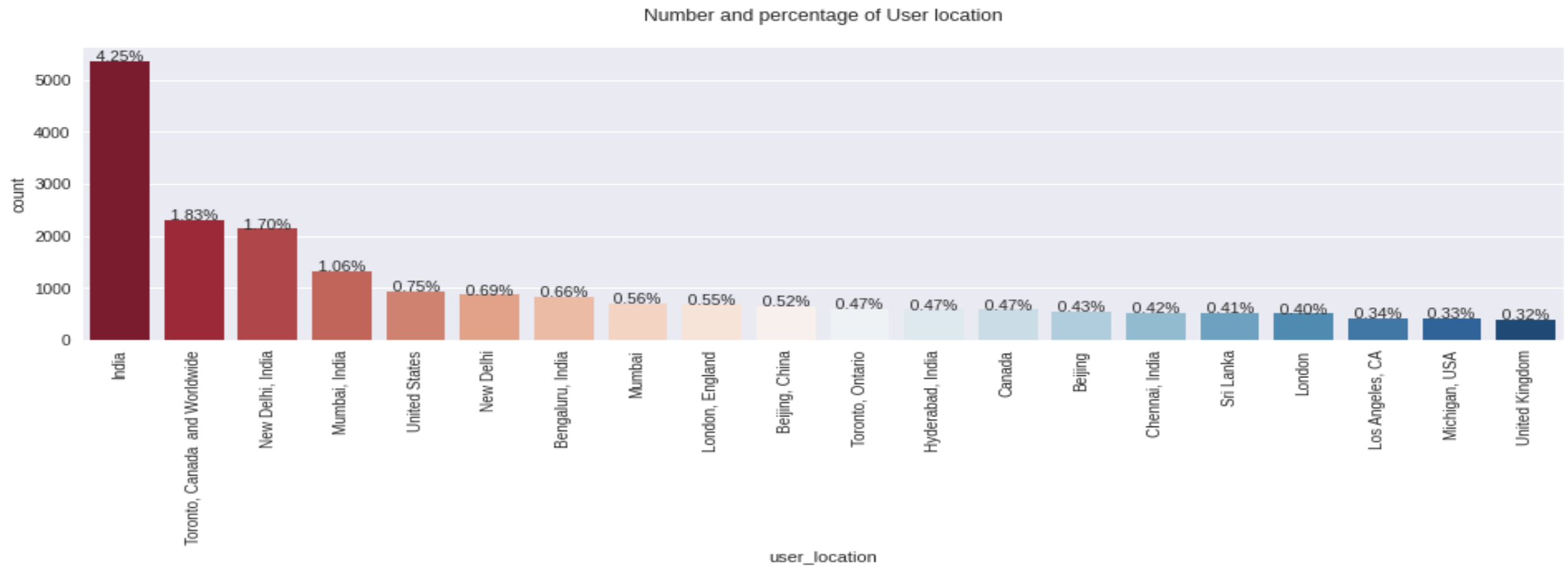

**Figure 9**: User Location of Tweets

### *3.3 Country wise 'Prevalent Word' Usage*

Here in collecting major utterances of vaccine related sentiments, the users are prioritized who have most of the tweets according to the location. So using word cloud, the most used terms in the USA, UK, Canada, and India are visualized in the figures below.

Fig. 10 shows the keywords used in USA like 'Moderna' which their own produced vaccine, 'Pfizer' almost gave similar accuracy like 'Moderna', then 'dose', 'shot', 'people', 'second dose' and such key terms showed their mixed feelings regarding vaccinations against COVID-19. Fig. 11 shows the most terms used in UK. ‘OxfordAstraZeneca’, ‘PfizerBioNTech’ such own produced vaccine names came in their tweets, besides ‘blood clot’, ‘feel’, ‘trial’ such alarming words found in their tweets. Fig. 12 and Fig. 13 shows the key terms used in the tweets of the people of Canada and India. Here ‘first dose’, ‘second dose’, ‘Moderna’, ‘Pfizer’, ‘Bharat BioNTech’, ‘death’, ‘emergency’ ‘Covishield’, clinical trial’ such words are very commonly used by the people of Canada and India.

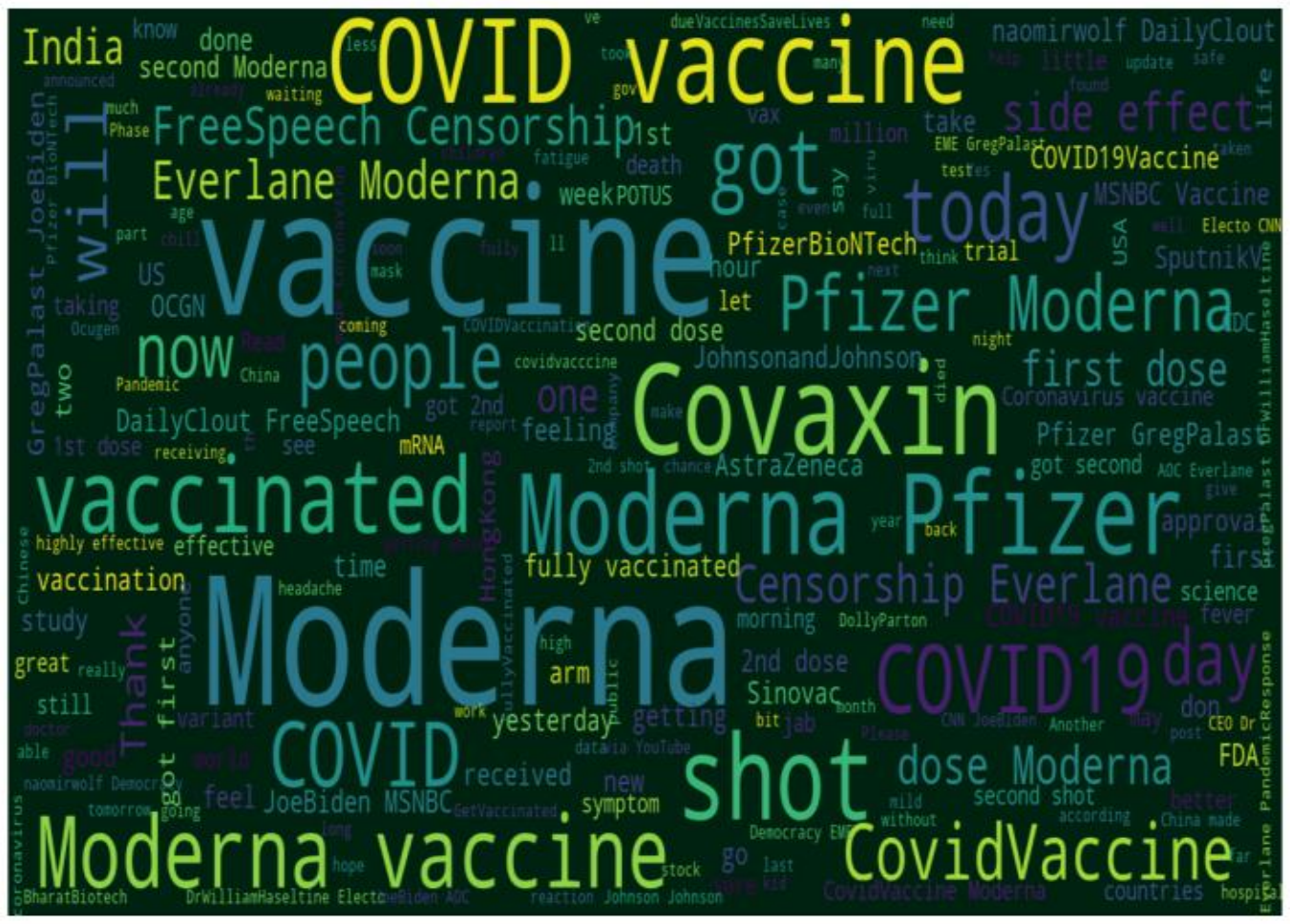

Prevalent words in tweets from US

**Figure 10**: Terms used most in US

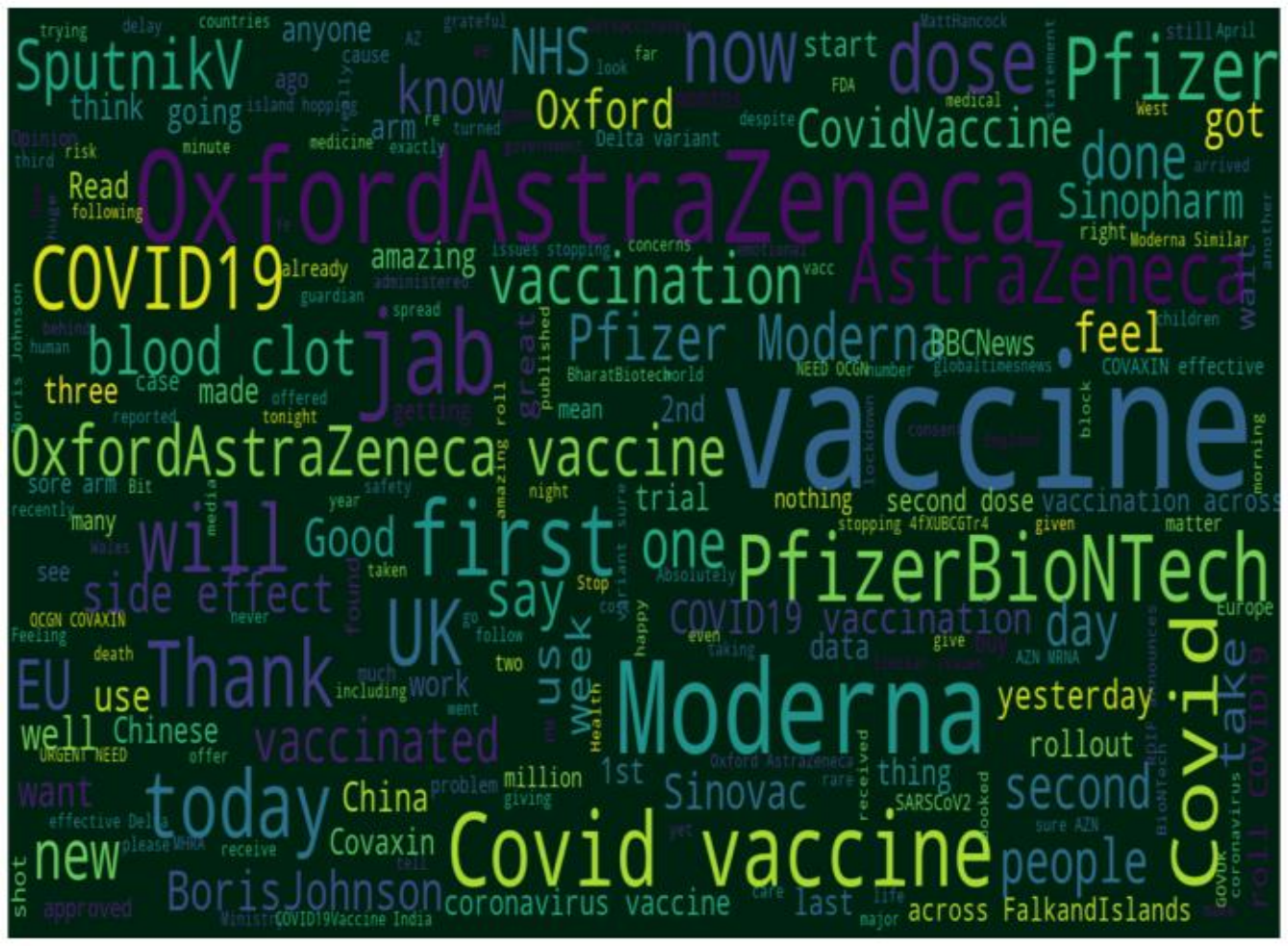

Prevalent words in tweets from UK

**Figure 11**: Terms used most in UK

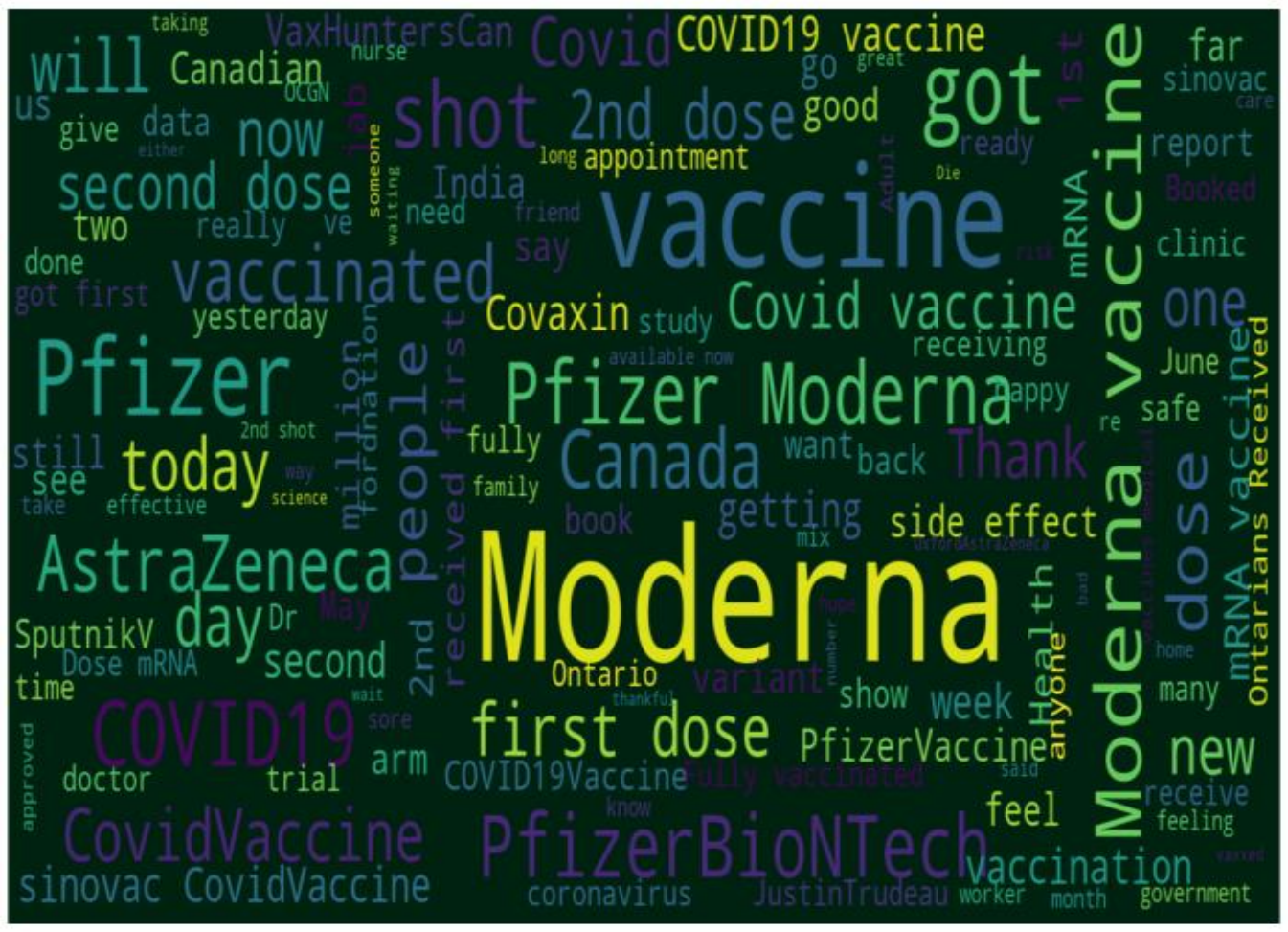

Prevalent words in tweets from Canada

**Figure 12**: Terms used most in Canada

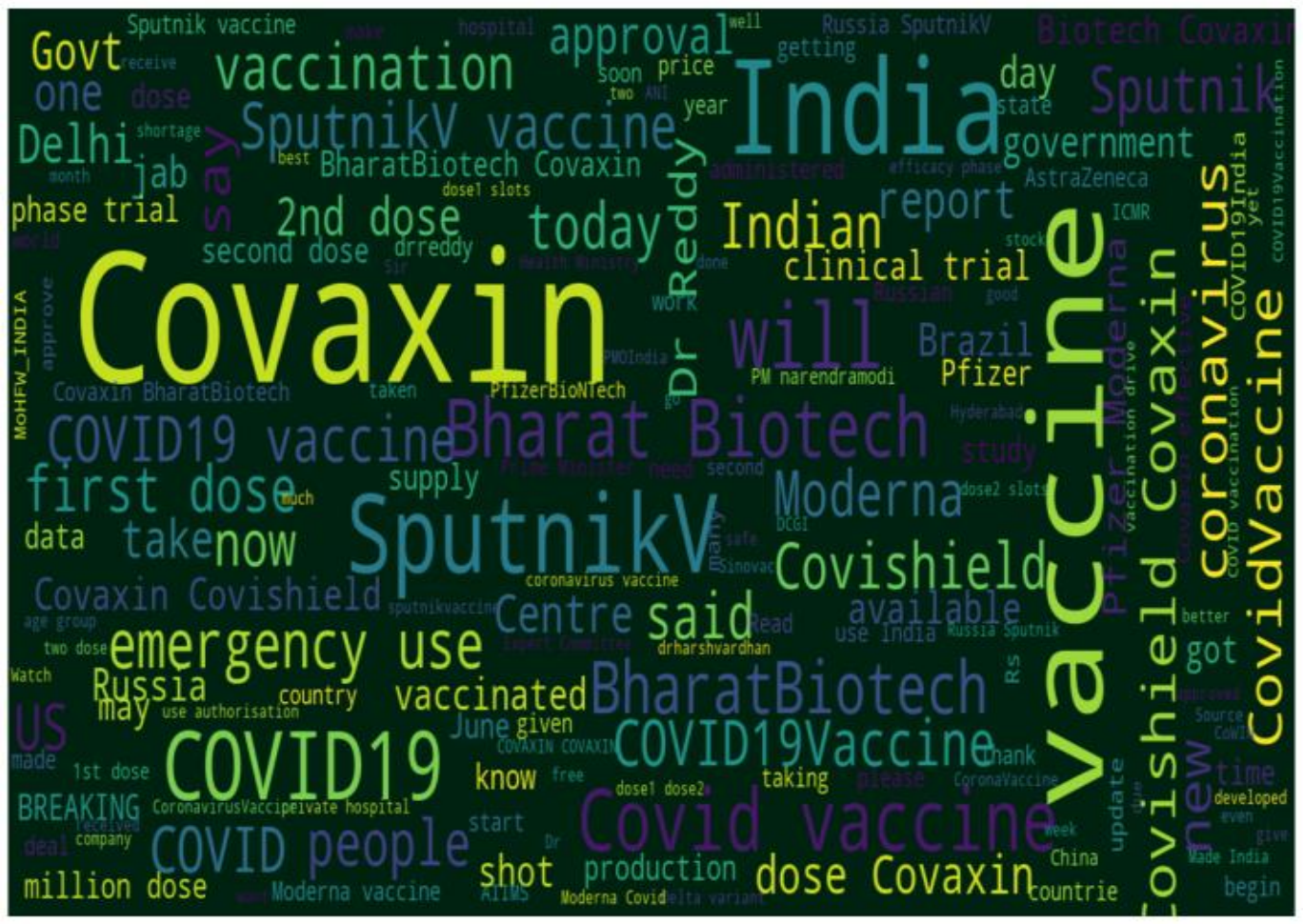

Prevalent words in tweets from India

**Figure 13**: Terms used most in India

### *3.4 Hashtag Counts Per Tweet*

Hashtags are important and highly used by maximum users in Twitter, and sometimes hashtags carry some significant meanings of particular events or trends. The total number of hashtags lies in the x-axis, and which hashtag is used most in the tweet is in the y-axis, which is called the Density Axis. The data scales down to a particular domain where the x-axis is numerical and the y-axis is decimal, as shown in Fig. 14.

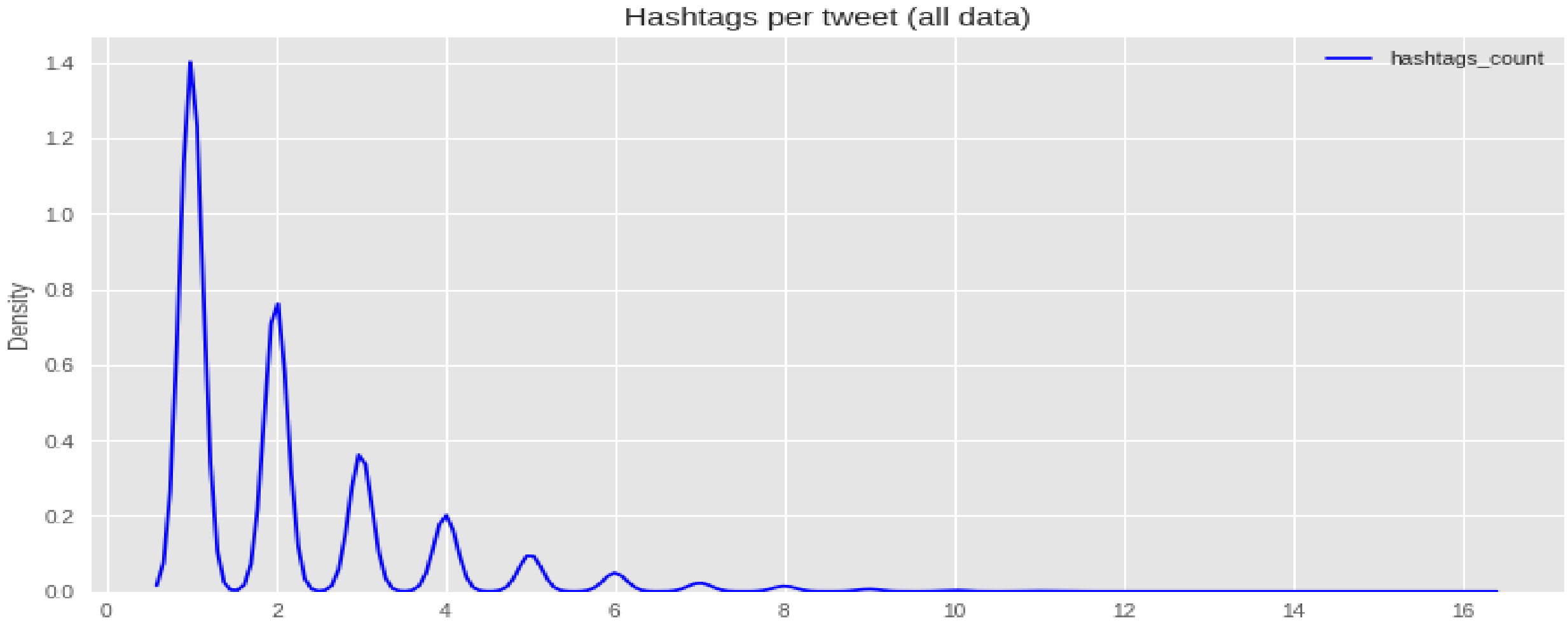

**Figure 14**: Hashtags per Tweet for All Data

### *3.5 Timeline of Tweet Reactions*

The number of tweets varies from time to time. Our research showed the changes in tweets during the start of the vaccination process until the latest time frame in Fig. 15.

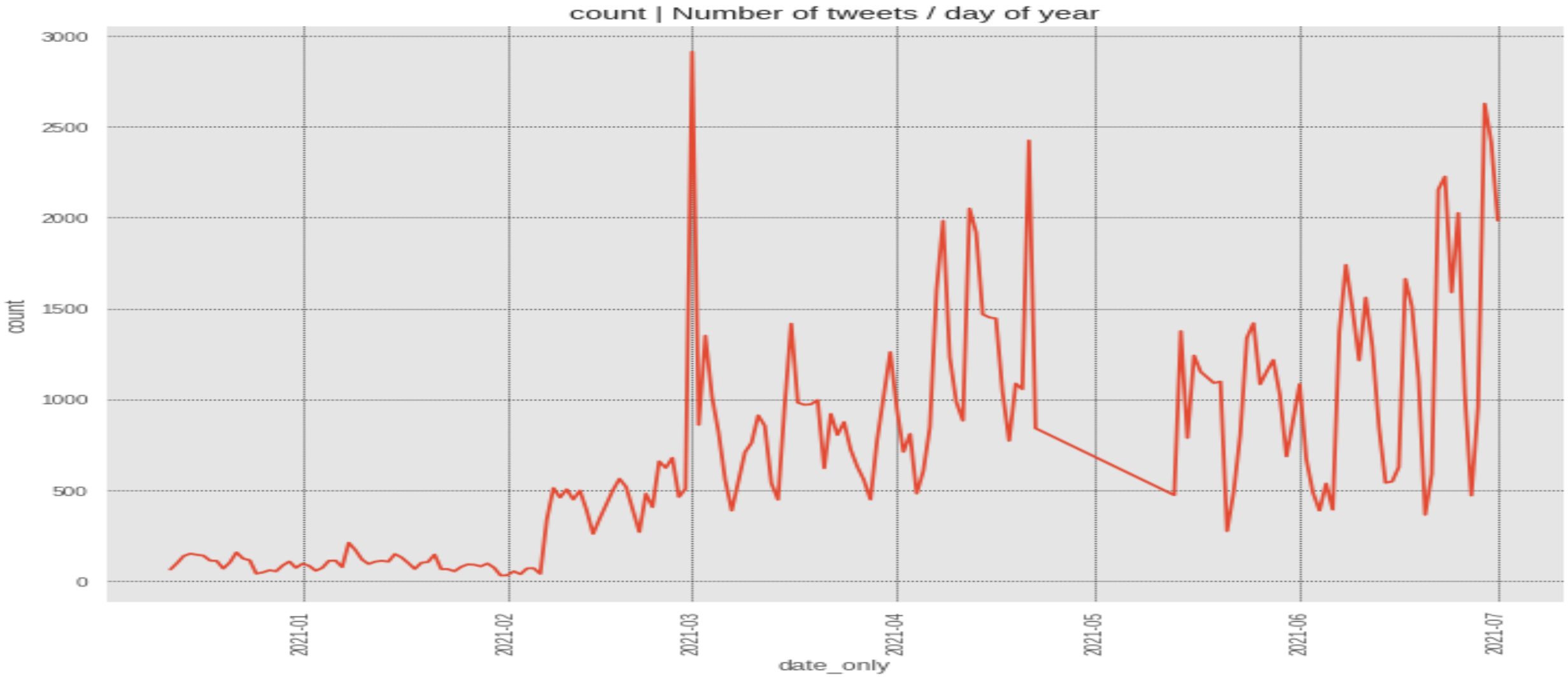

**Figure 15**: Day wise of Number of Tweets

From January’21 to the end of February’21 the tweets related to vaccines were less than 500; from March’21, it jumped to nearly 3000, indicating people were very excited about the vaccines after finishing the clinical trial and started implementing them, mass people. And from March’21 to till now, the tweets regarding covid-19 vaccines fluctuated from 1000 to 2500 per month, indicating the diverse emotions.

### *3.6 Sentiment Analysis and Evaluation*

This is one of the vital parts of our research, focusing on the Positive, Negative, and Neutral percentages to understand how emotions vary.

#### *3.6.1 Numbers and Percentage of Sentiment Criteria*

In Fig. 16, the numbers of tweets classified as Positive, Negative, And Neutral are demonstrated by Green bar charts, and percentages of these 3 categories are presented with Blue bar charts. For the charts colored Green, the x-axis determined the numbers of tweets, and the y-axis determined the sentiment classes. And in Blue charts, probability distributions are shown on the x-axis and in the sentiment classes on the y-axis.

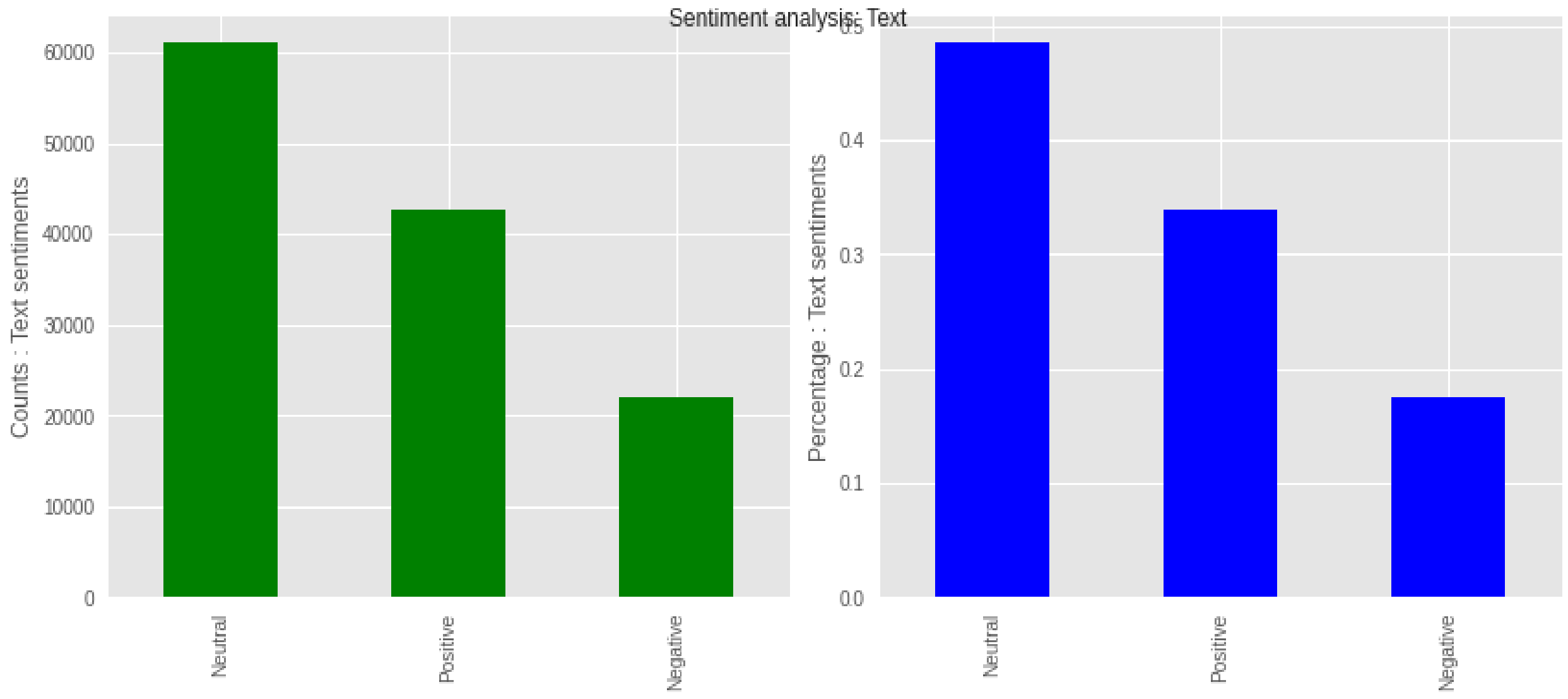

**Figure 16**: Numbers and Percentages of Sentiment Criteria.

From the datasets, there were 125,906 tweets that were analyzed through lexicon-based VADER and segregated into 3 parameters: Positive, Negative, and Neutral. Here, 42,765 numbers of Positive tweets with the percentage of 33.96% and 22,094 numbers of Negative tweets with the percentage of 17.55% is found from the analysis. There were 61047 numbers of Neutral tweets with a percentage of 48.49%. Neutral tweets are the majority here; negative reactions are also less, indicating that confusions, conflicts, and uncertainties related to COVID-19 vaccination procedures are still there. Optimistically, many people are still positive about these vaccines, whichmay motivate the vaccine producers and the rule-makers to control the Coronavirus.

#### *3.6.2 Timeline of these Sentiments*

As discussed earlier, sentiments or emotions evolve and transform with time. This research visualizes how the reactions change with time and how they fluctuate at different time intervals. Fig. 17 demonstrates how sentiment changes or shuffles as time goes by. 3 sentiment classes are shown in 3 colors. For positive, negative, and neutral, the colors used are green, red and blue, respectively.

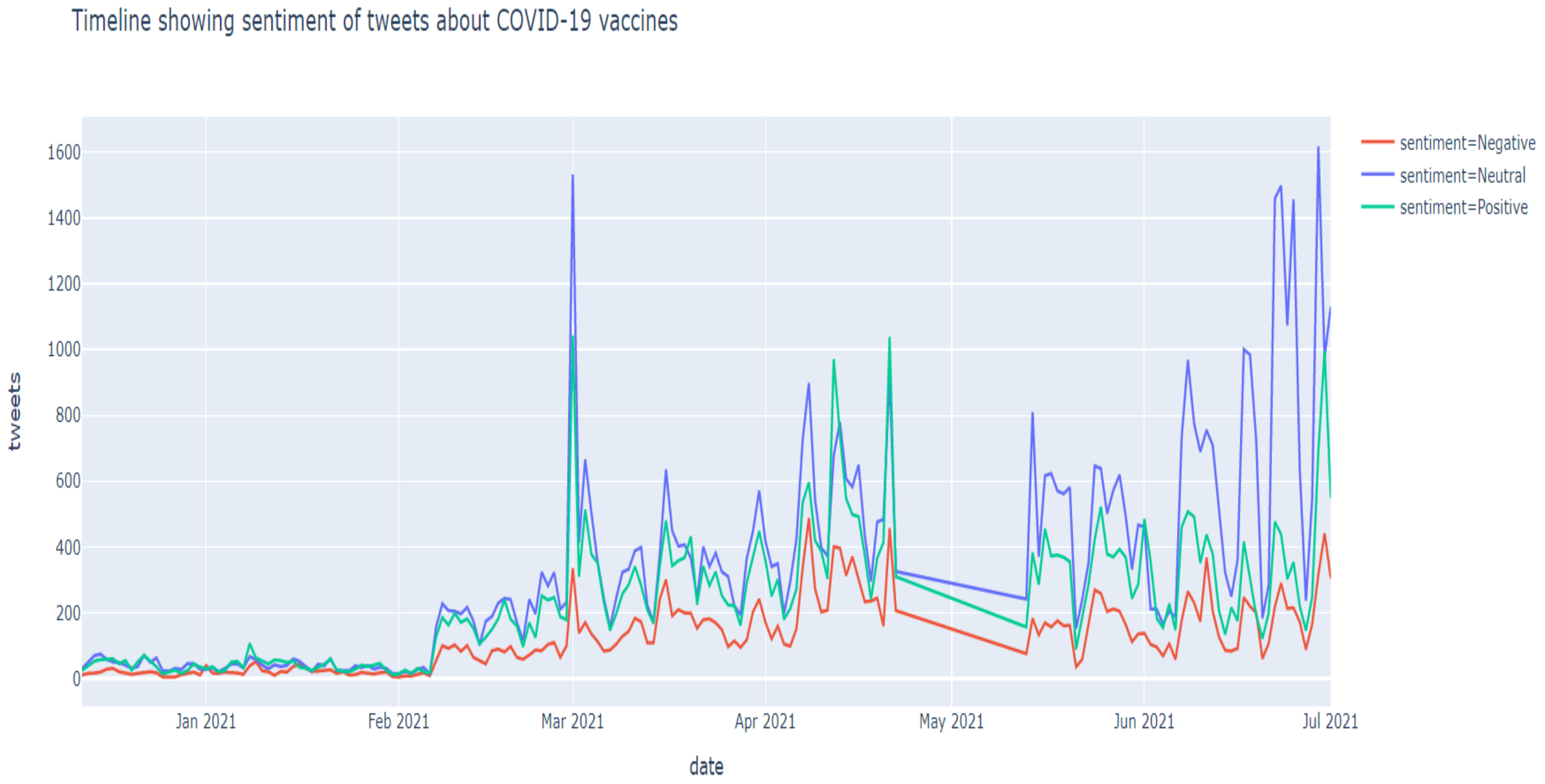

**Figure 17**: Sentiment Variation with Time

The fluctuation of emotions or sentiments with times during this pandemic is clearly visible above. From Fig. 17, it is clear that the sentiments were at the peak at the starting of March’21 when the final trial of vaccines ended. Neutral is the high polarity every month till July’21, and negative polarities have shown pretty less than Neutral in recent times.

*3.6.3 Sentiment Words according to Polarities*

Using WordCloud, the specific words or terms are classified into 3 polarity groups are shown below. In Fig. 18 (a), the terms used as Positive sentiments are showed, Fig. 18 (b) showed the Negative sentiment words, and Neutral sentiment words are shown in Fig. 18 (c).

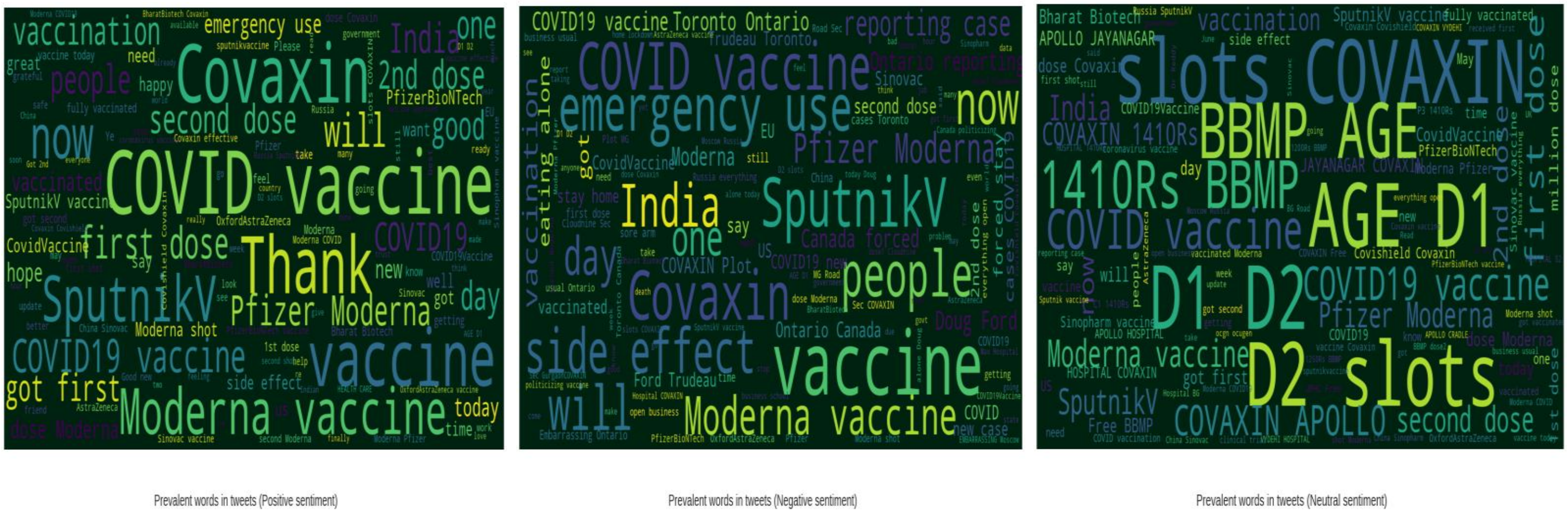

(a) (b) (c)

**Figure 18**: Words according to (a) Positive,(b) Negative,and (c) Neutral

### *3.7 Performance Evaluation*

For evaluating model performance, RNN based architectures LSTM and Bi-LSTM are implemented to check that they performed efficiently in our experiment. In the data sequencing and splitting part, we needed to convert processed data into vectors using tokenizing and transform values into target labels. The train test split methodfrom the scikit-learn library. Splitting data shapes arex_train: 94429, x_test: 31477, y_train: 94429,y_test: 31477. In this section, the performance metrics of both models and their prediction capabilities are presented.

#### *3.7.1 Performance Analysis with LSTM and Bi-LSTM*

To train our model with both LSTM and Bi-LSTM, used tools are listed in Table. 1 below.

**Table 1**: Training Parameters for LSTM and Bi-LSTM

| **Training factors** | **Components** |
|---|---|
| Platform | Google COLAB |
| GPU | Colab GPU (NVIDIA Tesla K80) |
| Optimizer | RMSProp |
| Loss | Categorical Cross-Entropy |
| Epoch | 10 |
| Activation | Softmax |

To train both of the models required some parameters. Parameters like loss function, optimizer, activation function and number epochs and batch size. Those parameters are very crucial for any model in order to train. In this work, 'Categorical Cross-entropy' is used as a loss function. This loss function helps to minimize errors for a particular class. RMSProp (Root Mean Square Propagation) optimizer was used to train both of these models. It's a well-known and extensively used optimizer. It helps to optimize the gradient descent during the training. Another essential feature is an activation function which is important to training a deep learning model. For this work, Softmax acts as an activation function. Softmax squashed down the main score gives a probability score for our final output. An epoch goes through all over the training data in a loop. To train LSTM and Bi-LSTM models on an all-over training dataset, we used only 10 epochs.

Fig. 19 and Fig. 20 exhibited the model accuracy and model loss for both training & test data of LSTM model. The accuracy of the validation set is 90.59% for the LSTM architecture.

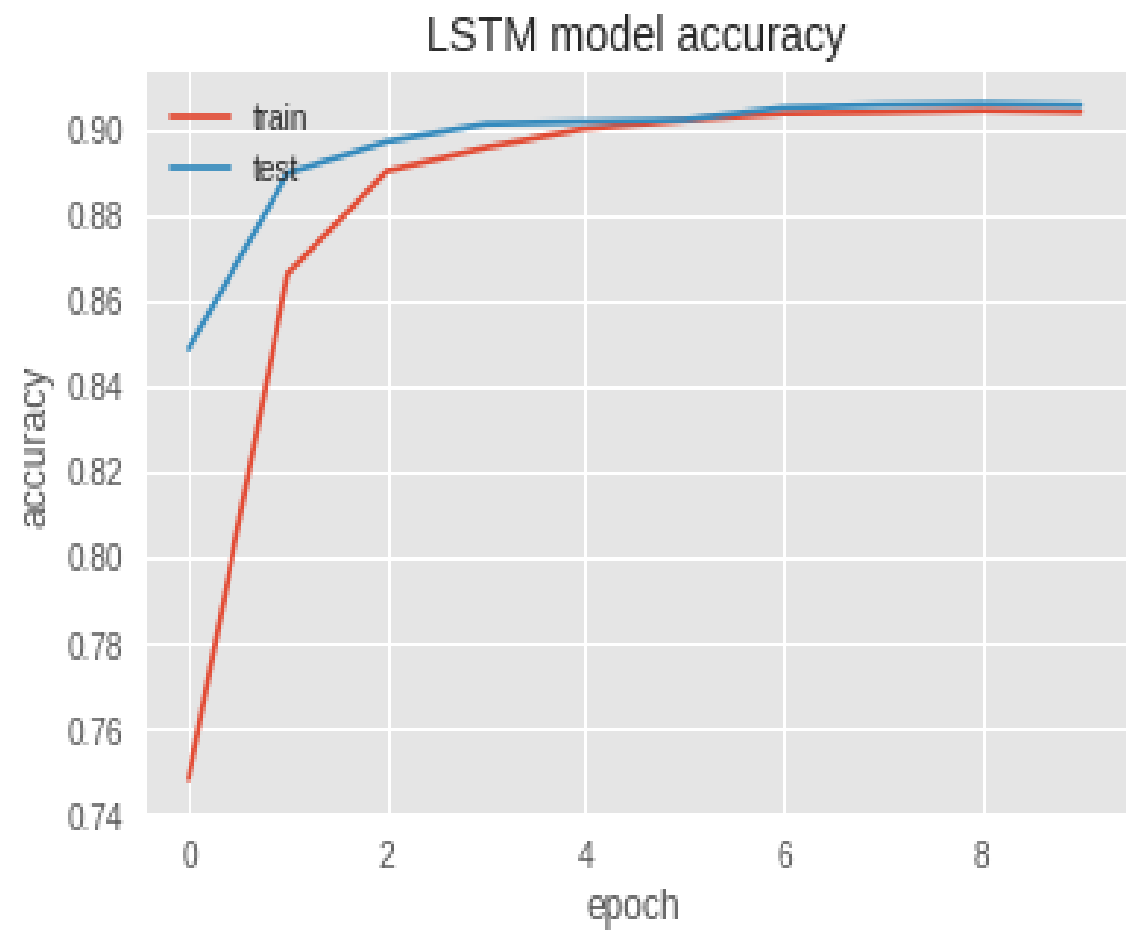

**Figure 19**: Accuracy of LSTM model

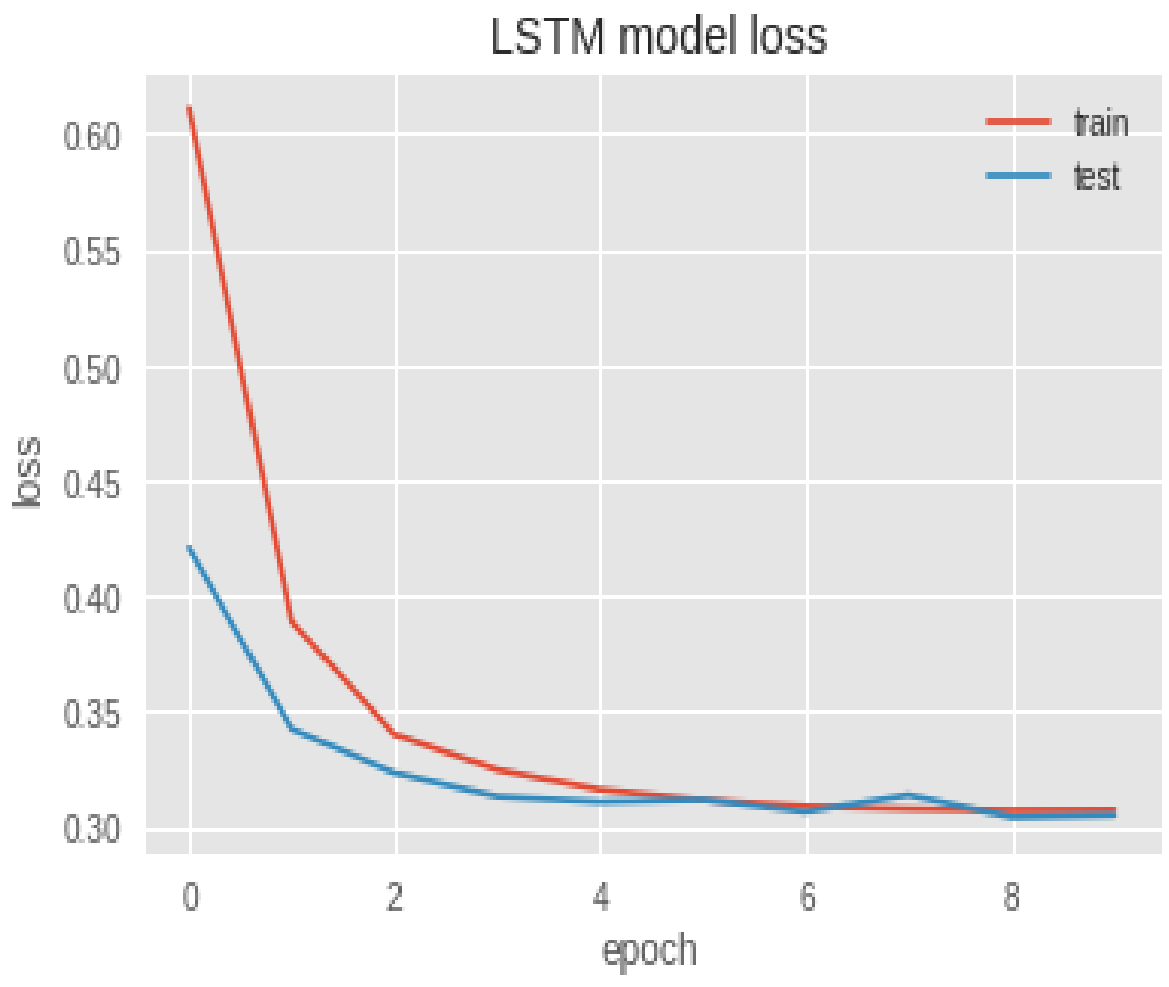

**Figure 20**: Loss of LSTM model

Model accuracy and model loss for both training and test data of Bi-LSTM model in Fig. 21 and Fig. 22. Here, for Bi-LSTM model, an accuracy of 90.83% found for both the validation set and the test set.

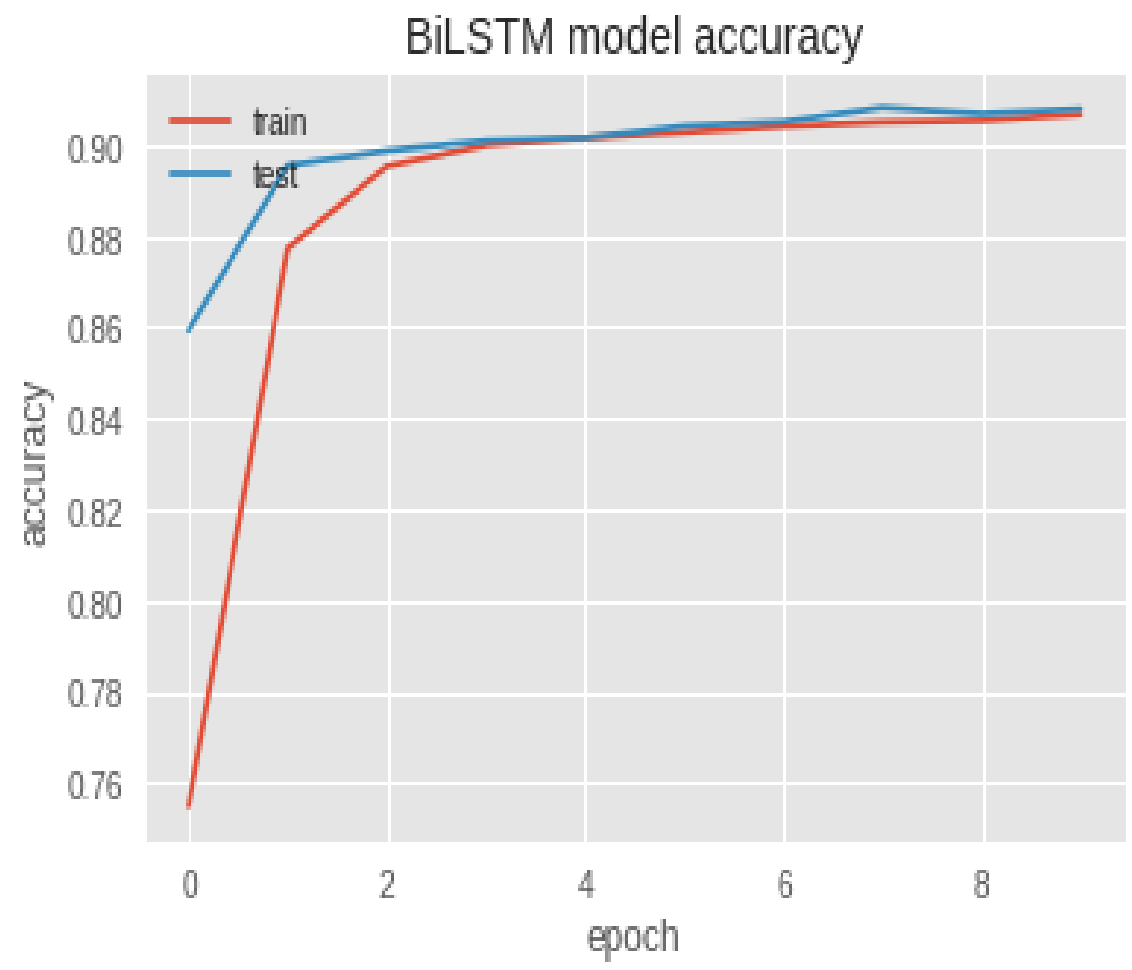

**Figure 21:** Accuracy of Bi-LSTM model

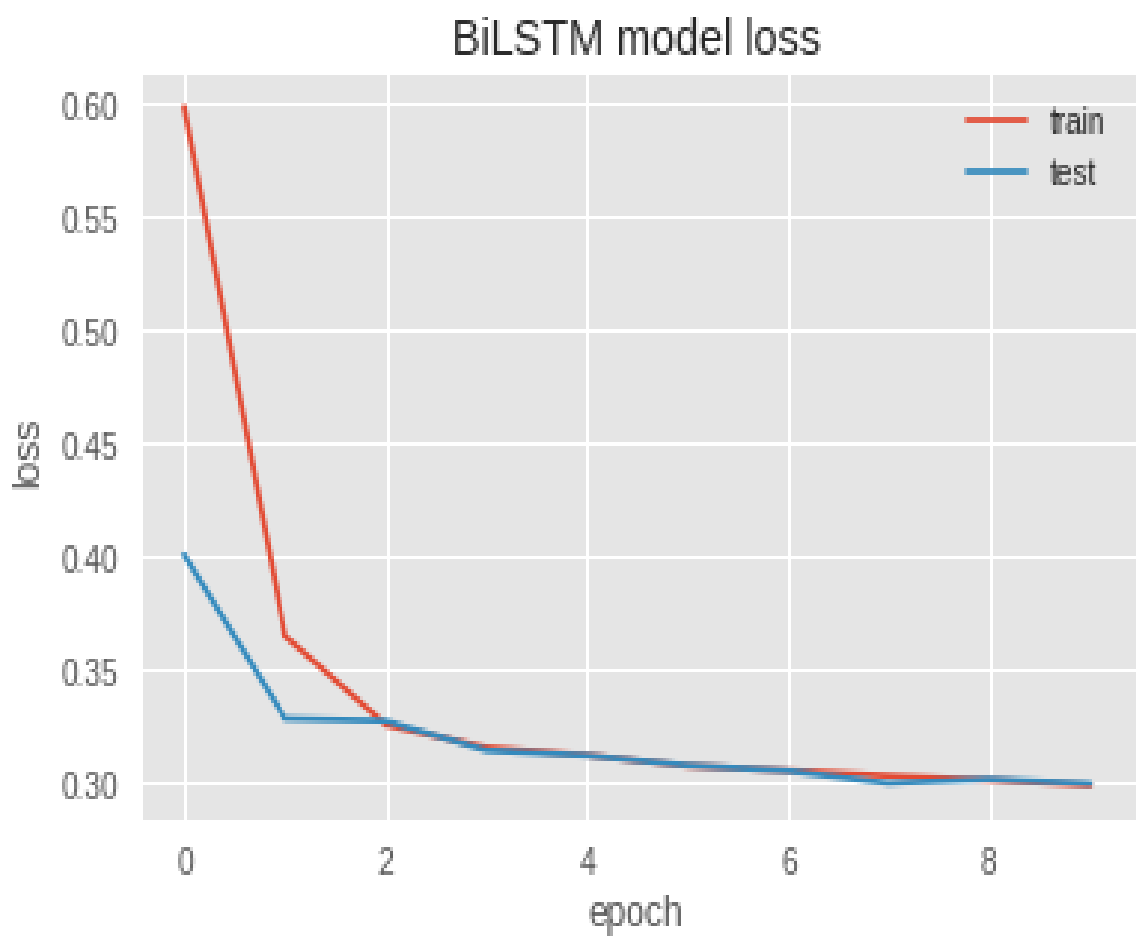

**Figure 22**: Loss of Bi-LSTM model

From the above Fig. 19 and Fig. 21, we can see that, in the LSTM and Bi-LSTM the training accuracy starts with near 75%. When evaluating fed data, LSTM gave 85% of accuracy at the beginning which ends with 90.59%, and Bi-LSTM showed accuracy of 86% at the beginning and ends with the accuracy of 90.83%. After implementing Bi-LSTM, accuracy of the model increased a little bit from 90.59% to 90.83%. By the time, increasing the number of epochs has helped to increase the accuracy of our both models, and the performance on the test set showed how effectively the performance of the model has been increased.

According to the architecture, LSTM only perform forward pass where Bi-LSTM perform both forward and backward passes. So when we fit our dataset to both models the LSTM takes the sequence only the forward way but Bi-LSTM takes the sequence in a way where every single text has its own sequence. This is a reason where network learn more deeply with Bi-directional LSTM thus gave a bit better performance in this sentiment analysis task.

### *3.7.2 Other Performance Metrics*

To evaluate the performance of the model based on different metrics, this work used the precision, recall, f1-score, and confusion matrix using different value rates like true positive (TP), false positive (FP), true negative (TN), false negative (FN).

*Precision:* It describes the performance of the model on the test data. It shows the number of models predicted correctly from all positive classes.

$$precision = \frac{TP}{TP + FP} \quad (6)$$

*Recall:* The percentage of total relevant results accurately classified by your algorithm is referred to as recall.

$$Recall = \frac{TP}{TP + FN} \quad (7)$$

*F1-Score:* The F1-score is simply the harmonic mean of precision and recall.

$$F1\ Score = 2 * \frac{Precision * Recall}{Precision + Recall} \quad (8)$$

Table. 2 shows the value of these performance metrics mentioned above.

**Table 2:** Outcomes of Performance Metrics

| | | Precision | Recall | F1-Score |
|---|---|---|---|---|
| **LSTM** | Class – 0 | 0.94 | 0.94 | 0.94 |
| | Class – 1 | 0.85 | 0.79 | 0.82 |
| | Class – 2 | 0.88 | 0.92 | 0.90 |
| | | Precision | Recall | F1-Score |
| **Bi-LSTM** | Class – 0 | 0.94 | 0.95 | 0.94 |
| | Class – 1 | 0.85 | 0.79 | 0.82 |
| | Class – 2 | 0.89 | 0.91 | 0.90 |

Neutral: Class - 0, Negative: Class - 1, Positive: Class -2.

Fig. 23 shows the confusion matrix for our prediction models for 3 different polarities.

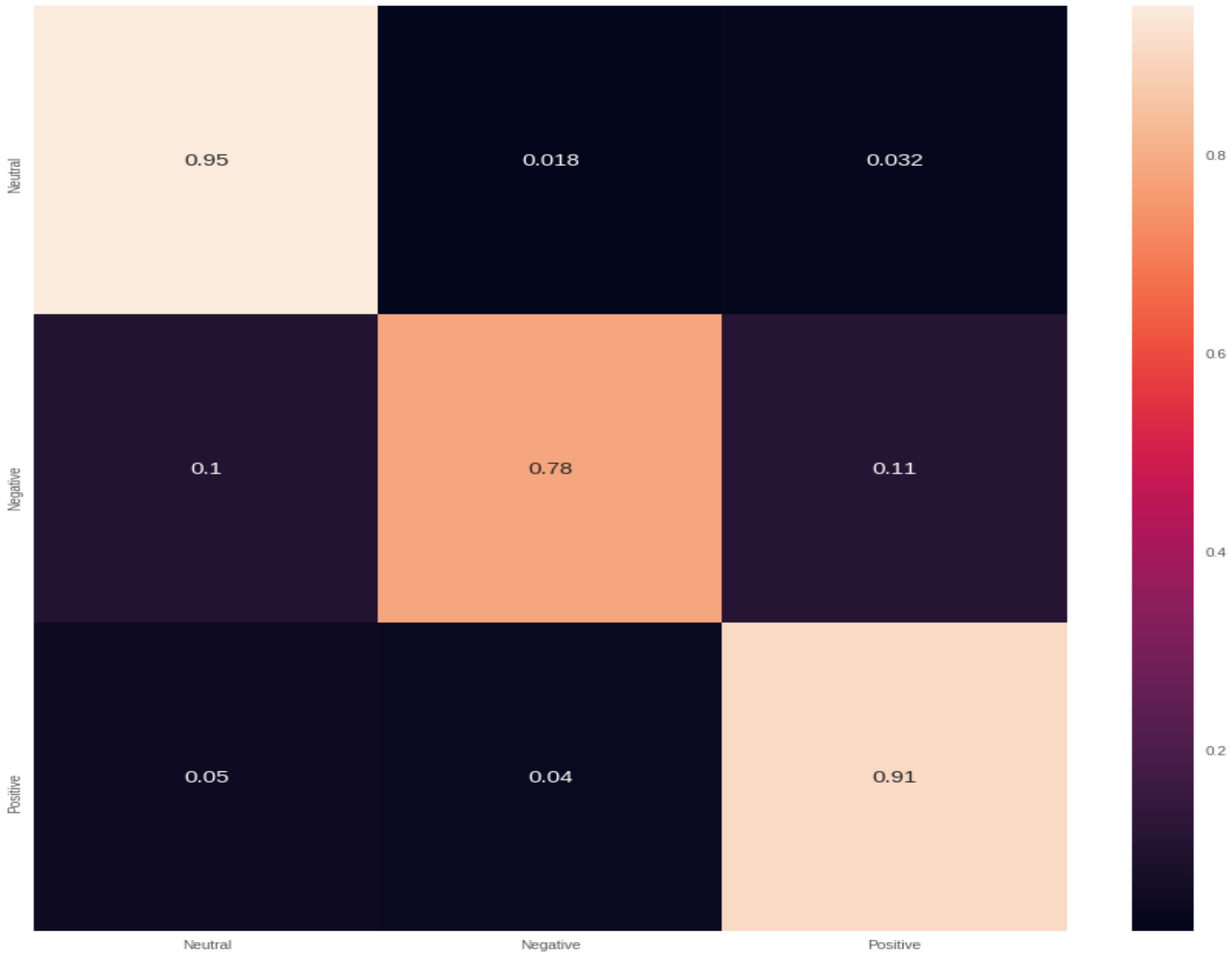

**Figure 23**: Confusion Matrix for Sentiments

And lastly, the model can correctly classify 91% of positive sentiments. From the confusion matrix, 94% is classified as Neutral which is True positive rate, where 73% accurately classify as Negative class. Besides, 2.6% and 3.2% being misclassified as Neutral where the highest wrong classifies in the Negative sentiment, which is about 11% and 15%. Eventually, the positive sentiment makes a good prediction, but about 5.4% and 3.9% wrongly classify as neutral and negative, respectively.

### *3.7.3 Sentiment Prediction Table*

The input from the tweets are randomly chosen and checked how well they can predict. Table.3 showed some outcomes of some refined tweets.

**Table 3**: Assessment of Sentiment Prediction

| | Text | Sentiment |
|---|---|---|
| 0 | Same folks said daikon paste could treat a cytokine storm #PfizerBioNTech | Positive |
| 1 | While the world has been on the wrong side of history this | Negative |
| | year hopefully the biggest vaccination effort weveev | |
| 2 | #coronavirus #SputnikV #AstraZeneca #PfizerBioNTech #Moderna #Covid19 Russian vaccine is created to last 2 4 years | Positive |
| 3 | Facts are immutable, Senator, even when you’re not ethically sturdy enough to acknowledge them 1 You were born i | Neutral |
| 4 | Does anyone have any useful advice/guidance for whether the COVID vaccine is safe whilst breastfeeding | Neutral |
| 5 | it is a bit sad to claim the fame for success of #vaccination on patriotic competition between USA Canada UK and | Positive |
| 6 | There have not been many bright days in 2020 but here are some of the best | Positive |
| 7 | Covid vaccine You getting it #CovidVaccine #covid19 #PfizerBioNTech #Moderna | Neutral |
| 8 | #CovidVaccine States will start getting #COVID19Vaccine Monday #US says #pakustv #NYC #Healthcare #GlobalGoals | Neutral |

## 4 Conclusion

Our research shows how 'Deep Learning' techniques are used in 'Sentiment Analysis' tasks. Basic NLP based tools are implemented here to understand the sentiment in 3 possible polarities like Positive, Negative, Neutral and our findings showed the 33.96% of people are positive, 17.55% people are negative and 48.49% people are neutral till the month July of 2021 in response of the vaccination procedure going all across the globe. RNN based LSTM and Bi-LSTM are also incorporated in our research to determine how accurately and precisely our built models can predict and analyze the sentiments. LSTM architecture showed 90.59% accuracy and Bi-LSTM model shows 90.83% accuracy, and both models showed a good prediction score in precision, recall, f-1 scores and confusion matrix calculation. Thus many people took the decision to vaccinate themselves, and a good number of people are still confused. Many of them are frightened and many of them directly refused to be vaccinated. Our research will help them to make decisions and understand the global situations and opinions of the people from different parts of the world.

This research will help the health researchers to get the proper knowledge of the issues regarding the vaccination process. The companies who produce vaccines, Governments or Health Ministries of different countries, or policymakers of this sector like World Health Organisation (WHO) [21] or others, can have a proper idea about whether their vaccine is effective or not, and the percentage of this effectiveness. They can understand that in which sector they have to improve so that people can have faith in this vaccination process. Moreover, they may have a clear idea about how many doses they should be ready with according to the wishes of the mass people. In Health Science Research, our research will be an added benefit to finding out the proper scenario as well as pros and cons of the vaccination process of Covid-19. We believe we will be a little but effective part to help the upfront fighters against this Novel Coronavirus and keep our lives healthy and safe.

**Data Availability Statement:** The data utilized to support this research findings is accessible online at: https://www.kaggle.com/gpreda/all-covid19-vaccines-tweets

**Acknowledgment:** The authors are thankful for the support from Taif University Researchers Supporting Project (TURSP-2020/211), Taif University, Taif, Saudi Arabia.

**Conflicts of Interest:** The authors declare that they have no conflicts of interest to report regarding the present study.